%% file: neurips_2025.tex
\documentclass{article}

\PassOptionsToPackage{numbers, compress}{natbib}

\usepackage[final]{neurips_2025}

\usepackage{multirow}

\usepackage[utf8]{inputenc} 
\usepackage[T1]{fontenc}    
\usepackage{hyperref}       
\usepackage{url}            
\usepackage{booktabs}       
\usepackage{amsfonts}       
\usepackage{nicefrac}       
\usepackage{microtype}      
\usepackage[table]{xcolor}         
\definecolor{mynangrey}{rgb}{0.5, 0.5, 0.5}
\usepackage[pdftex]{graphicx}
\usepackage{subcaption}
\usepackage{amsmath}

\usepackage{booktabs}
\usepackage{adjustbox}
\usepackage{geometry}
\usepackage{float}
\usepackage{comment}

\usepackage{siunitx}
\usepackage{longtable}
\usepackage{arydshln}

\newcommand\doubleplus{\mathbin{+\mkern-10mu+}}

\newcommand{\ours}{ConTextTab}

\newcommand{\custompar}[1]{\noindent\textbf{#1:\;}}

\definecolor{mygrey}{rgb}{0.75, 0.75, 0.75}
\definecolor{mynangrey}{rgb}{0.85, 0.85, 0.85}
\newcommand{\NAN}{\textcolor{mynangrey}{N/A}}

\title{{ConTextTab}: A Semantics-Aware Tabular In-Context Learner}

\author{%
Marco Spinaci\textsuperscript{\normalfont$*\textrm{1}$}\quad%
Marek Polewczyk\textsuperscript{\normalfont$*\textrm{2}$}\quad%
Maximilian Schambach\textsuperscript{\normalfont$*\textrm{2}$}\quad%
Sam Thelin\textsuperscript{\normalfont$\textrm{2}$}\\
\textsuperscript{\normalfont$\textrm{1}$}SAP France \quad \textsuperscript{\normalfont$\textrm{2}$}SAP SE\\
\texttt{\{firstname.lastname\}@sap.com}
}

\begin{document}

\maketitle
\phantomsection\def\thefootnote{*}\footnotetext{Equal contribution.}\def\thefootnote{\arabic{footnote}}

\begin{abstract}
Tabular in-context learning (ICL) has recently achieved state-of-the-art (SOTA) performance on several tabular prediction tasks. Previously restricted to classification problems on small tables, recent advances such as TabPFN~\citep{tabpfnv2} and TabICL~\citep{tabicl} have extended its use to larger datasets. Although current table-native ICL architectures are architecturally efficient and well-adapted to tabular data structures, their exclusive training on synthetic data limits their ability to fully leverage the rich semantics and world knowledge contained in real-world tabular data. At the other end of the spectrum, tabular ICL models based on pretrained large language models such as TabuLa-8B~\citep{tabula8b} integrate deep semantic understanding and world knowledge but are only able to make use of a small amount of context due to inherent architectural limitations. With the aim to combine the best of both these worlds, we introduce \textbf{ConTextTab}, integrating semantic understanding and alignment into a table-native ICL framework. By employing specialized embeddings for different data modalities and by training on large-scale real-world tabular data, our model is competitive with SOTA across a broad set of benchmarks while setting a new standard on the semantically rich CARTE benchmark. Code and model checkpoints are available at: \href{https://github.com/SAP-samples/sap-rpt-1-oss}{https://github.com/SAP-samples/sap-rpt-1-oss}.
\end{abstract}

\section{Introduction}

Tables with information spread across rows and columns remain a predominant data format in many real-world applications~\citep{chui2018notes}, making their understanding through machine learning algorithms critical. Despite the great success of deep learning approaches in natural language processing and computer vision, leveraging large amounts of pretraining data, conventional machine learning methods such as gradient boosted trees and their variants remain the predominant state-of-the-art (SOTA) across tabular prediction benchmarks. Recently, however, applying the in-context learning (ICL) paradigm to tabular prediction tasks has shown promising results by enabling pretraining of deep learning models across a large set of heterogeneous tables, constituting a new SOTA on small to medium tabular prediction tasks~\citep{tabpfnv2}. In this setting, predicting a target value $y$ based on the input features $\boldsymbol{x}$ of a row in a table $T$ additionally uses further rows from $T$ (the context), including their target values. This enables the model to adapt to new, unseen prediction problems at inference time, removing the need for task-specific fine-tuning.

This approach was pioneered by the transformer-based TabPFN~\citep{tabpfnv1}. Pretrained on large amounts of synthetically generated classification tasks, its latest incarnation TabPFNv2~\citep{tabpfnv2} produces SOTA results on tabular datasets with up to \num{10000} samples for both classification and regression tasks. In recent work, TabICL~\citep{tabicl} extends the success story of this approach to even larger datasets, using special tabular embedding modules improving on the quadratic scaling in both the number of features and rows present in the TabPFN architecture.

Common to both TabPFN and TabICL is that they are trained entirely on synthetically generated numerical data, with categorical features produced by an indexing procedure. While using synthetic data has many advantages, in particular its diversity at scale, a consequence is that the data does not contain any semantically meaningful values as found in real-world applications, both in the form of column names, and categorical or free-text entries. Furthermore, such data contains no additional data types such as times or dates that are abundant in practice. Consequently, these models do not utilize such information in a semantically meaningful way even when it is present at inference. In particular, column names are not used in either TabPFN or TabICL, and categorical features are encoded via one-hot or ordinal encoding disregarding any underlying semantics. We argue that semantic understanding can be captured by training on a large number of real-world tabular datasets. A primary example of this philosophy is TabuLa-8B~\citep{tabula8b}, turning the pretrained large-language model (LLM) Llama 3-8B~\citep{dubey2024llamareducedauthors} into a tabular ICL model by fine-tuning it on around 3 million tables of the T4 dataset proposed therein. However, utilizing pretrained LLMs for tabular tasks has several limitations: Most importantly, textual serialization and tokenization of the input table is not token efficient, effectively limiting the maximum context length that can be processed. For example, TabuLa-8B operates on a maximum of 32 context rows. Furthermore, the tokenization schema and autoregressive nature of LLMs are not adapted to the tabular structure, resulting in a linear non-uniform token sequence, as cell values can be tokenized into different amounts of tokens, losing the 2D structure of the underlying data. Finally, the serialization and autoregressive processing result in an architecture that is neither row nor column permutation invariant -- a property often desirable for tabular data~\citep{towardsfoundation}.

Aiming to bridge these approaches, we propose a table-native ICL model trained on the real-world T4 dataset~\citep{tabula8b}, using embeddings tailored to different data modalities, in particular incorporating semantic embeddings of column names and categorical values. The resulting model is competitive to existing table-native ICL approaches across a range of tabular prediction benchmarks (OpenML-CC18~\citep{openmlcc18}, OpenML-CTR23~\citep{openmlctr23}, TALENT~\citep{talent}, and TabReD~\citep{tabred}), and achieves a new SOTA for the semantically rich CARTE benchmark~\citep{CARTE}, in particular in the low-data regime.

\section{Related Work}
\custompar{Tabular deep learning} Prediction on tabular data has traditionally been dominated by decision tree algorithms, particularly boosted variants like XGBoost~\citep{xgboost}, LightGBM~\citep{lightgbm}, and CatBoost~\citep{catboost}. These models deliver strong performance but require separate training for each dataset and cannot leverage pretraining. Hence, much work has been done to transfer the success of deep learning methods and general pretraining to the tabular setting. Early examples include the FT-Transformer~\citep{ft-transformer} and Xtab~\cite{xtab}, whereas more recent approaches have shown consistently good performance overtaking boosted trees, for example TabR~\citep{tabr}, RealMLP~\citep{realmlp}, CARTE~\cite{CARTE}, TabM~\citep{tabm}, or ModernNCA~\citep{modernnca}.

\custompar{In-context learning on tabular data} TabPFN~\citep{tabpfnv1} broke the long-standing dominance of boosted trees on small classification tasks, outperforming them by using row-level ICL. Pretrained on a large amount of synthetic tabular data, it generalizes to new tasks at inference time without fine-tuning or hyperparameter optimization. A recent variant, TabDPT~\citep{tabdpt}, showed that equally excellent results can be achieved by training on real-world data using similarity-based retrieval for the context examples -- an idea previously investigated in the TabR approach~\citep{tabr} -- and further unlocked regression in this setting. Generalizing the row-based encoding, cell-based ICL introduced with TabPFNv2\footnote{In the following, we will focus solely on TabPFNv2 and refer to it simply as TabPFN.}~\citep{tabpfnv2} and utilized also by TabICL~\citep{tabicl} extended this success to larger datasets with up to \num{10000} and more samples, even outperforming SOTA AutoML solutions such as AutoGluon~\citep{autogluon} on certain benchmarks.   

\custompar{Semantics and real data} Capturing the rich semantics of real-world tabular datasets is a desirable property of a tabular foundation model, enabling the transfer of world knowledge across prediction tasks in addition to statistical patterns. The CARTE~\citep{CARTE} architecture enables pretraining across a variety of real-world sources while capturing table semantics. It achieves SOTA results on the semantically rich CARTE benchmark, although it requires task-specific fine-tuning.

Modern LLMs have both deep semantic understanding and extensive world knowledge. Several works approach tabular ICL by tuning LLMs on tabular tasks, e.g.\ TabLLM~\citep{tabllm}, LIFT~\citep{lift}, or TabuLa-8B~\citep{tabula8b}. In particular, the works by \citet{tabula8b} are note-worthy for curating the T4 dataset, containing roughly 3\,M tables extracted and processed from the TabLib collection~\citep{tablib} and for its excellent results in the very low data-regime. 

\section{Method}

\begin{figure}
\centering
\vspace{-3mm}
\includegraphics[width=0.8\linewidth]{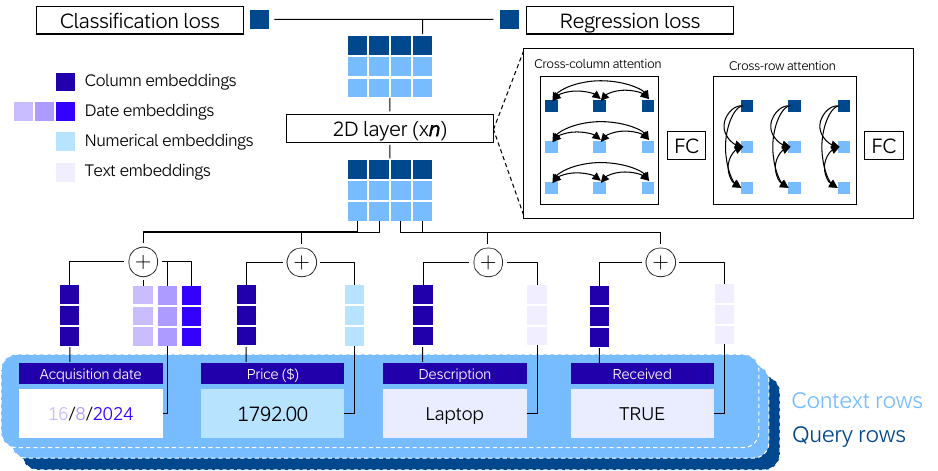}
\caption{Our proposed model architecture illustrating the integration of data type-specific embeddings, an interleaved attention backbone, and customized output heads.}
\label{fig:architecture}
\end{figure}


To overcome the aforementioned limitations of existing table-native ICL methods and bridge the gap to LLM-based ones, we propose \ours{}, a semantics-aware table-native ICL model. To this end, we perform several key modifications to the TabPFN architecture and utilize large-scale pretraining on real-world data. An overview of our proposed architecture is given in Figure~\ref{fig:architecture}.

\subsection{Encoding}\label{sec:encoding}

We encode data differently depending on its modality -- i.e.\ text, date, or numeric type. Column headers are also encoded, playing the role of positional encodings as used in TabPFN or TabICL. 

\custompar{Text} We transform each text cell to an embedding vector using a pretrained text embedding model. Note that we apply this to both free text columns as well as categorical columns, which can then retain the meaning in their labels. Any off-the-shelf embedding model can be used for this purpose -- e.g.\ many are available based on the BERT architecture~\citep{devlin-etal-2019-bert}. As many values have to be embedded for each table, there is a natural trade-off between accuracy and speed. We settle for a comparably small and fast model such as \texttt{all-MiniLM-L6-v2} \cite{all-MiniLM-L6-v2,wang2020minilm}, since the amount of semantic meaning it captures is already much richer than conventional categorical encodings, and we defer to Section~\ref{sec:ablation} for an experimental ablation. Since the embeddings are potentially of a different dimension than the target embedding dimension $d$, we apply a learnable linear layer to the text embeddings.

\custompar{Date} In order to endow our model with knowledge of both the relative meaning of a dates (e.g.\ being able to compare two dates) and special dates (e.g.\ recurring holidays), we embed each of the numbers representing day, month, and year separately and sum the three resulting vectors. In particular, this is more token efficient than encoding dates into multiple features as is common in established preprocessors such as the one used by AutoGluon.

\custompar{Numerical} As numbers do not contain semantic meaning beyond their value, we apply a one-dimensional encoding. To make this procedure more robust during training, first, we clip columns between the 2\% and 98\% quantiles of the distribution. Second, we scale them to have zero mean and unit variance as is common. By Chebyshev's inequality, this bounds resulting values to the interval $(-7.1, 7.1)$, thus avoiding exploding gradients during training. Finally, the resulting number is multiplied by a learnable vector and a bias is added. If the original value was NaN, 0 is used instead, so the bias works as an ``is-NaN'' flag. An alternative embedding scheme is described in \ref{sec:alternatives}.

\custompar{Column headers} We embed column headers with the same model used for text cells. The result is passed through a separate learnable linear layer to map to the correct target dimension and summed with the cell embedding. 

After summation, the embeddings are normalized via layer normalization. Note that all the above embeddings are fully equivariant under permutations of either rows or columns. This equivariance makes the predictions more reproducible and robust and eliminates some of the need (such as the mapping of category to ID, or of column to ID) for bagging in models such as TabPFN. Semantic embeddings of cell values have been previously investigated in other works~\cite{yak2023ingestables,ye2024crosstable,CARTE,PORTAL}, however, details of how the information is consumed differ in each implementation.

\subsection{Backbone}
We leave the TabPFN architecture mostly unchanged, with alternating ``horizontal'' (cross-column) and ``vertical'' (cross-row) self-attention transformer layers. In cross-column attention, each row is considered as a different batch element, and conversely. Following TabPFN, cross-column attention has no masking, while cross-row attention is masked so that each row can only attend to the provided context. To increase the modularity of our code, the feedforward MLP block of the transformer encoder is repeated after each self-attention block so that ``horizontal'' and ``vertical'' blocks have the same structure.
For sizing, we use the nomenclature popularized by BERT, e.g.\ calling ``base'' the variant with 12 layers and a hidden dimension of 768. However, the real number of (non-embedding) weights is twice that of a BERT model, due to the presence of both ``horizontal'' and ``vertical'' layers.

Model weights can be optionally shared between each instance of the transformer block, consisting of two interleaved attention layers. Such an architecture can be interpreted as a recurrent neural network, unrolled in depth rather than in time. Such an iterative structure results in parameter efficiency and the possibility of stacking more blocks in comparison to the traditional approach. Empirically, we observed that sharing weights did not affect model performance and thus we use weight sharing as the default option for our model in the following.

\subsection{Decoding}\label{sec:decoding}
\custompar{Classification} We apply a standard cross-entropy loss after an MLP with at least as large an output dimension as the number of classes. This, however, imposes two limitations. First, the number of classes seen in pretraining cannot be exceeded at inference, without resorting to suboptimal schemas such as hierarchical classification, as used in TabPFN's many-class extension, TabDPT, or TabICL. Secondly, it prevents us from using the semantic value of the class label. In fact, for the model to know what, say, class ``0'' means, it needs to build an a priori knowledge of that class ID. Therefore, we must create an additional special input encoding, only for the target column, in addition to the ones described in Section~\ref{sec:encoding}. Even though this breaks equivariance under permutation, we retain it as we found it to be effective in the most common scenario of few-classes classification.

\custompar{Regression} The model predicts the floating-point value of the target, clipped and normalized as described in Section~\ref{sec:encoding}. Empirically, we have found this simple schema to work well. During training, an $L_2$ loss is applied, and during inference, only the inverse transformation of normalization is applied to the prediction.

\subsection{Alternative architectures} \label{sec:alternatives}


\custompar{Number encoding and decoding via binning}
\noindent{\emph{Encoding -- soft binning.}} We split the numbers into bins defined via quantiles to ensure uniform distribution. In order not to lose any fine-grained information while keeping the number of bins limited, we use soft binning: each of the, say $n$, bins is associated with a quantile $\{q_i\}_{1, \dots, n}$ of order $\frac{2i-1}{2n}$. Then, any point $x \in [q_1, q_n]$ is encoded via the linear combination $\lambda v_i + (1-\lambda)v_{i+1}$, where $x = \lambda q_i + (1-\lambda) q_{i+1}$, $0 \leq \lambda \leq 1$, and $v_1, \dots, v_n$ are learnable vectors. Anything outside that interval is mapped to either $v_1$ or $v_n$.

\noindent{\emph{Decoding -- bin averaging.}} The regression is converted to a classification task for the bin it belongs to. During training, cross-entropy loss is applied with hard labels. At prediction time, this provides a probability distribution across all bins, $(p_1, \dots, p_n)$; the prediction is then given by $p_1 q_1 + \dots + p_nq_n$.

\custompar{Supervised clustering head}
We introduce an alternative method to perform classification that has the advantage of retaining semantic meaning not only for features but also for labels. As further benefits, full equivariance is preserved, without the need to map classes to IDs, and the limitation on the number of classes supported for prediction is lifted, improving upon the constraint currently present in many tabular ICL approaches such as TabPFN, TabDPT, or TabICL. For this, we take inspiration from \citet{clustertabnet}: For each row, the final output corresponding to the target column is mapped by a two-layer MLP to a vector $x$. We then build a matrix of shape $(n_{\mathrm{query}}, n_{\mathrm{context}})$ by computing the cosine similarities between these vectors $x_i$, $x_j$ for each query row $i$ and context row $j$. During training, this is compared against the adjacency matrix with value 1 if two rows belong to the same class, and 0 otherwise: We compute the element-wise binary cross-entropy loss between this adjacency matrix's entries and the clipped cosine similarities. That is, if $x_i$ is the vector from above for row $i$ and $c_i$ is its class, then the loss is given by 
\begin{equation}
\text{Loss} = -\sum_{\substack{i \in \mathrm{query} \\ j \in \mathrm{context}}} \log ( s_{ij}) \delta_{c_i=c_j} + \log(1-s_{ij}) \delta_{c_i \neq c_j}, \quad s_{ij} = \text{clip}\Big(\frac{x_i \cdot x_j}{\|x_i\|\|x_j\|}, \varepsilon, 1-\varepsilon \Big)
\end{equation}
including a small margin $\varepsilon > 0$ for numerical stability.
In this way, rows in the same class are pushed to similar embeddings, while rows in different classes can be either opposite or orthogonal. This enables the use of the same semantic-preserving text encoding for the target class of context rows as used for column names and input features, while also supporting an arbitrary number of target classes. A possible use case for this scenario is a situation with many target classes with very few rows per class, allowing extraction of information from semantically similar target classes.

\custompar{Induced Set Attention Blocks}
One of the main limitations of the standard multi-head attention is its quadratic complexity. While memory can be effectively compressed to linear~\cite{efficientattention,flashattention}, that is not the case for runtime. This poses a significant challenge when dealing with large tables. Taking inspiration from \cite{isab, perceiver, tabicl}, we experiment with methods to handle larger context more efficiently.

Our proposed architecture replaces some of the multihead attention blocks within TabPFN with ISAB blocks from~\cite{isab}. Adopting the notation from~\cite{isab}, let $\mathrm{MAB}(X, Y)$ denote the standard multihead attention block from the transformer architecture, where $X$ represents the queries and $Y$ serves as both the keys and values. This is followed by a feedforward block, a skip connection, layer normalization, and ultimately an intermediate block. Let $I$ be so-called inducing points~\cite{isab}, an additional set of parameters of shape $(n, d)$ for an arbitrary $n$, with $d$ denoting the hidden dimension, and define
\begin{equation}
    \mathrm{ISAB}(C \doubleplus Q) = \mathrm{MAB}\big(C \doubleplus Q, \mathrm{MAB}(I \doubleplus Q, C \doubleplus Q)\big),
\end{equation}
where $C$ and $Q$ denote context and query rows, respectively, and $\doubleplus$ denotes concatenation. In each attention computation, we apply masking so that only context and inducing vectors are attended to. For a detailed overview of the architecture, refer to Appendix Figure~\ref{fig:ISAB_diagram}.

Since tables can have a very large number of rows but rarely exceed a few hundred columns, we apply the ISAB block only to cross-row attention. In our architecture, we only use the ISAB block for the first $m$ blocks (e.g., $m = 3$), followed by one $\textnormal{MAB}(I \doubleplus Q, C \doubleplus Q)$ block, and finally use standard attention for the following layers. This both improves training stability and reduces inference time because the corresponding attention has a bounded number of tokens: $\textnormal{MAB}(I \doubleplus Q, I \doubleplus Q).$


\section{Experimental Setup}
\subsection{Training and inference}\label{sec:training}

For pretraining, we use the T4 dataset~\citep{tabula8b}. We discard tables with fewer than 150 rows, which leaves \num{2.18}\,M tables with a median of 750 rows and 9 columns. We randomly select 1000 rows, then between 50 and 900 rows as query, and use the rest as context. Subsequently, we randomly select one target column, excluding all date columns, numerical columns with more than 50\% NaN values, and other columns having more than 20\% of unique values. Finally, we up-sample non-numeric columns to have roughly the same proportion of regression and classification tasks. We train each model for between 4 and 10 million steps (i.e., 2 to 5 epochs) until convergence. We use a micro batch size of 1 and accumulate gradients to simulate a batch size of 256 (or 128 for smaller models of ``mini'' size). To improve stability, we employ gradient clipping and the AdamW optimizer with a maximum learning rate of $10^{-4}$, reached after a linear warm-up phase of 1000 gradient updates. Under this setup, we train a ``base'' sized model on a single H100 GPU, reaching a throughput of roughly 10 tables/s. Hence, full training takes between 4 and 12 days depending on the number of steps. Using the default parameters $n=12$, $d=768$, $d_{\textrm{ff}}=3072$ results in a total of 172\,M parameters which is reduced to 16\,M trainable parameters when using weight sharing.



We also experimented with curriculum learning, by further adding in a second step using the same training data as \citet{tabdpt}. This data has fewer tables, 123, but many more cells in each table with a median of 11\,k rows and 34 columns. In this second step, we increased the number of rows used for training to 4000. We refer to Section~\ref{sec:ablation} for an analysis of the impact of training data size on model performance, which we found to be crucial.

At inference, we apply 8-fold bagging, similar to the ensembling used by TabPFN. That is, from the original train split of a given evaluation dataset, we sample 8 times $c$ context rows with replacement, make a separate prediction with each collection using the same model, and average the predictions (regression values or classification probabilities). While, during the default training, $c$ never surpasses 950, we can scale this during evaluation as much as memory permits. The combination with bagging allows \ours{} to use up to $8c$ points as context. By default, we use a context size of $c=8192$ at inference. To limit runtime, in each bag, we sample up to 500 columns if the dataset has more.

\subsection{Evaluation}\label{sec:setup_evaluation}

\custompar{Datasets} We use a variety of tabular prediction datasets to evaluate and compare our approach to established baselines and other SOTA methods. Namely, we evaluate all models on the following benchmarks: OpenML-CC18~\citep{openmlcc18}, a pure classification benchmark; OpenML-CTR23~\citep{openmlctr23}, a pure regression benchmark; TALENT~\citep{talent}, a recently introduced diverse benchmark containing over 300 classification and regression benchmarks. Here, we focus on a subset containing 45 datasets that are representative of the overall performance of the baselines investigated in the original works, which we refer to as the TALENT-Tiny benchmark; TabReD~\citep{tabred}, a small but challenging benchmark of large datasets representative of practical prediction tasks; and finally CARTE~\citep{CARTE}, a mixed classification and regression benchmark containing highly semantic features and few numerical ones.

Across all benchmarks, we evaluate 91 regression and 112 classification tasks, ranging from 5 to 3\,k columns and 400 to roughly 400\,k training examples. Due to the large number of evaluated datasets, we do \emph{not} perform cross-validation but evaluate a fixed test split for each task. For models that do not explicitly use a validation split, we concatenate the train and validation splits.
We refer to Appendix~\ref{app:dataset} for a visualization of the dataset statistics and further details.

\custompar{Baselines} We compare our approach to several established classical methods as well as recent ICL and other deep learning models.
In particular, we compare with TabPFN~\citep{tabpfnv2}, TabICL~\citep{tabicl}, and TabDPT~\citep{tabdpt} as the most recent table-native ICL methods. For all pretrained models, we use the latest available release and checkpoints as of July 2025.
As the SOTA on the CARTE benchmark, we also evaluate the CARTE~\cite{CARTE}. As additional well-performing recent deep learning approaches, we evaluate RealMLP~\citep{realmlp} and TabM~\cite{tabm}. 
Furthermore, we evaluate SOTA boosted tree baselines, namely XGBoost~\citep{xgboost}, LightBGM~\citep{lightgbm}, and CatBoost~\citep{catboost}.
The tree baselines as well as RealMLP and TabM are evaluated in the default (D) or meta-tuned default variants (TD)~\citep{realmlp}, as well as via Parzen-tree estimator-based hyperparameter optimization, either with a fixed holdout set (HPO) or ensembled via 5-fold inner cross-validation with per-split HPO (HPO, CV). For this, we use the implementation provided by \texttt{pytabkit}~\cite{realmlp}.
Additionally, we evaluate several common, non-tuned baselines from \texttt{scikit-learn}~\citep{sklearn}, namely naive, linear, and KNN estimators, as well as a random forest and histogram-based gradient boosting tree estimators.
Finally, as the gold standard in tabular prediction, we evaluate AutoGluon~\citep{autogluon}, an AutoML solution stacking and ensembling many of the baselines outlined above. 
We refer to Appendix~\ref{app:baselines} for full details on the used baseline versions and parameters as well as details about data preprocessing and encoding.

\custompar{Metrics} Throughout, we show (mean) accuracy for classification tasks as well as (mean) $R^2$ score for regression tasks. Since we have observed some severe outliers in some regression tasks that greatly affected average scores, we perform a soft-clipping. That is, all negative $R^2$ scores are mapped to $[-1, 0)$ via $\tanh$ while keeping positive scores untouched. This retains the relative ordering of scores while being smoothly differentiable and less distorting to the mean.

As averaging across a large number of datasets with varying performance can blur relative performance across models, we also evaluate (mean) rank. Within each benchmark, we calculate the mean rank across all constituent datasets and models. Full rank performance is averaged across all datasets across all benchmarks, not as per-benchmark averages as some benchmarks contain only few tables. To reduce the impact of small noise, we count model performances as ties when their evaluation scores lie within 0.005 of each other. 
Furthermore, we report critical difference (CD) diagrams, win ratio matrices, and p-values enabling 1 vs 1 model comparisons in the Appendix.

\section{Results}

\begin{figure}[H]
\centering
\begin{subfigure}[t]{0.55\textwidth}
    \centering
     \includegraphics[width=\linewidth,clip,trim={0 7mm 0 8mm}]{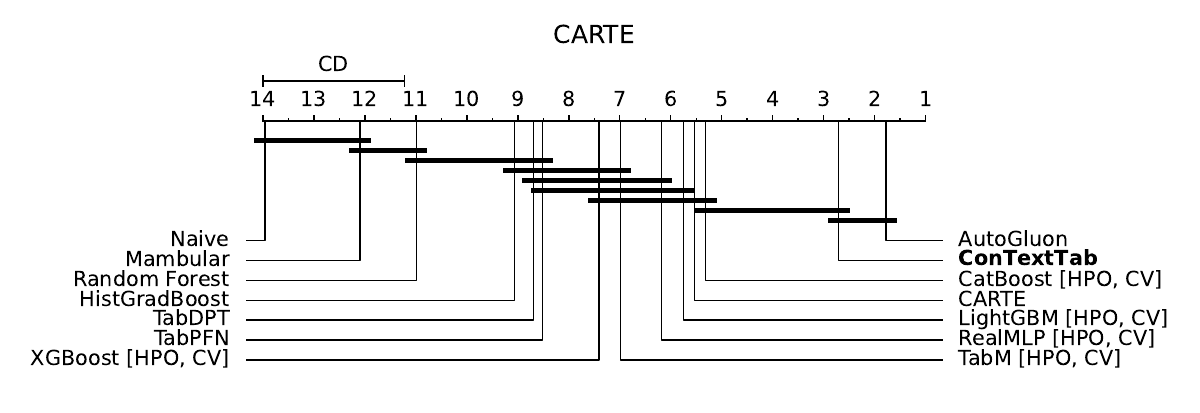}
    \label{fig:cd_diagram}
\end{subfigure}
\hfill
\begin{subfigure}[t]{0.44\textwidth}
    \centering
    \includegraphics[width=\linewidth]{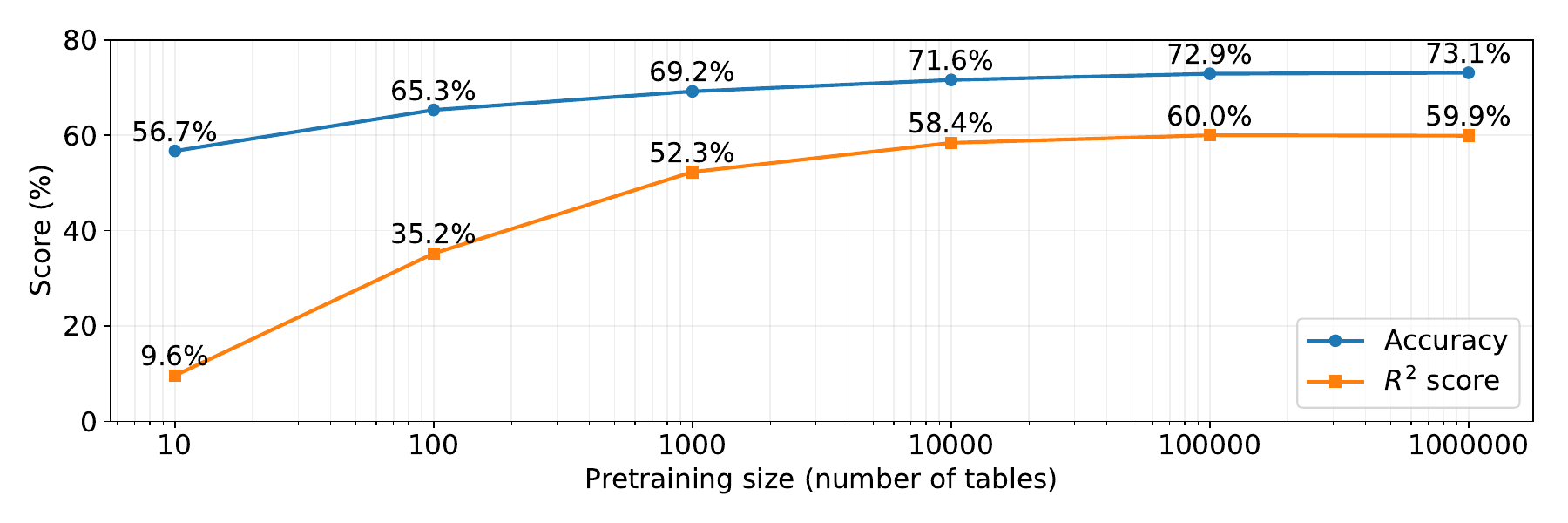}
    \label{fig:pretraining_size}
\end{subfigure}
\caption{Left: critical difference diagram between \ours{} and several baselines, across the CARTE benchmark. Right: Impact of pretraining dataset size on validation accuracy and $R^2$ scores.}
\label{fig:critical_difference_and_pretraining_size}
\end{figure}

The main results, evaluating \ours{} and baselines, are shown in Table~\ref{table:main_results}. Here, we exclude AutoGluon from the overall rank and best-model comparison as it is ensembling and stacking a multitude of model architectures, making it difficult to highlight architectural strengths and weaknesses. Overall, our model performs competitive across all non-semantic benchmarks while setting a new standard on the semantically rich CARTE benchmark, where it ranks best. In fact, this is statistically significant over all models but CatBoost and AutoGluon, both of which are extensively tuned and ensembled, as shown in the critical difference diagram for CARTE in Figure~\ref{fig:critical_difference_and_pretraining_size} (left). Notably, \ours{} is significantly better than other tabular ICL approaches, which highlights the importance of incorporating semantic understanding. On the other hand, TabPFN, not incorporating semantics, performs worse than tuned as well as untuned trees on CARTE. For details, we refer to Appendix~\ref{app:additional-results}.

On the other end, the absolute performance of \ours{} for non-semantic benchmarks, while good, falls behind tuned and ensembled boosted trees as well as RealMLP, which achieves SOTA in terms of ranking across all evaluated datasets (albeit at a much higher training time and cost).
Nevertheless, these differences in ranking are mostly not statistically significant. To this end we report per-benchmark CD diagrams as well as win-ratio and p-values for detailed model comparison in Appendix~\ref{app:additional-results-extended}. Namely, \ours{} does not perform significantly worse than the best-ranking model on all non-semantic benchmarks excluding OpenML-CTR23 (where it is however not significantly different to tuned ensembled trees).

Hence, \ours{} sets a new standard on semantically rich datasets while performing on-par to existing approaches on non-semantic ones. Nevertheless, the performance of AutoGluon generally indicates further headroom to improve single-model performance and tabular ICL in particular.

\begin{figure}
  \centering
  \includegraphics[width=\linewidth]{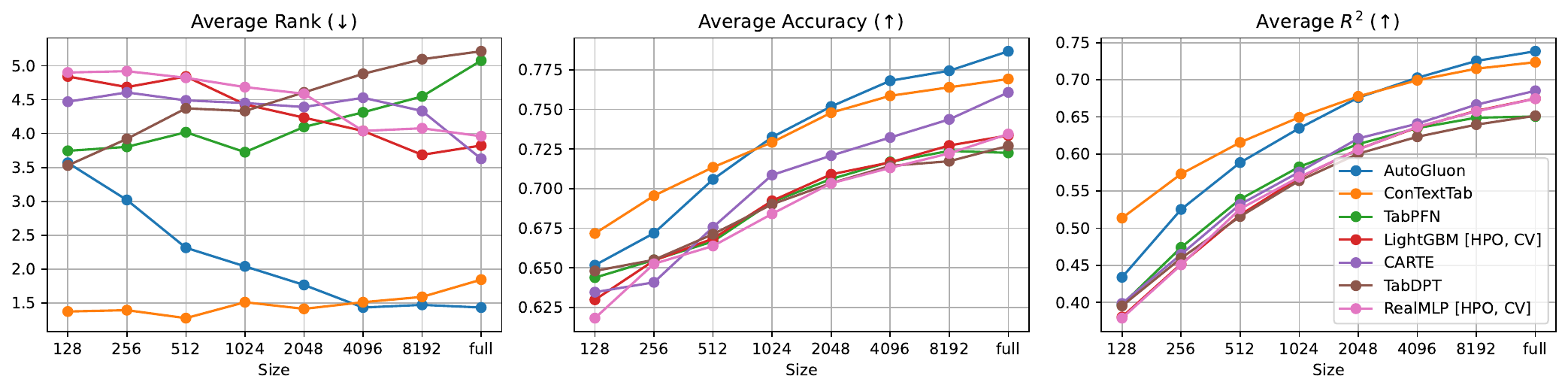}
  \caption{Average rank, accuracy, and regression results on the CARTE benchmark across various data subsets, ranging from 128 rows to the full size.}
  \label{fig:carte-sub}
\end{figure}

Finally, tuned trees perform particularly well for large datasets as represented by the TabReD benchmark, even outperforming AutoGluon in some instances. Motivated by this, we investigate the dependence on dataset size more closely and the performance on the CARTE benchmark across varying subsampled sizes, ranging from 128 rows to the full dataset, in Figure~\ref{fig:carte-sub}. \ours{} consistently outperforms other models across all sample sizes and even surpasses AutoGluon for up to 2048 training samples. This highlights the strong capabilities of tabular ICL but also the need for further research into scaling these architectures to effectively deal with much larger context sizes as well as training datasets.
We investigate the dataset-size dependent rank performance across all 203 datasets in Appendix~\ref{app:additional-results-data-size}, further highlighting these challenges.

Additional results, including additional baselines, CD diagrams, as well as 1 vs 1 win ratios and p-values, are provided in Appendix~\ref{app:additional-results-extended}. 
Furthermore, we refer to Appendix~\ref{app:runtime-analysis} for an analysis of the model runtime in comparison to other baselines.



\setlength{\arrayrulewidth}{1.2pt}  
\renewcommand{\arraystretch}{1.1}
\begin{table}
\caption{Performance comparison across all evaluated benchmarks, depicting mean accuracy (Acc) for classification and (soft-clipped) $R^2$ score for regression tasks, in percent. Missing values, due to architectural limitations or failed evaluations, are denoted as N/A and excluded from the rank calculations. Models are sorted according to their ranking on the CARTE benchmark.}\vspace{0.5mm}
\label{table:main_results}
\centering
\begin{adjustbox}{max width=\textwidth}
\input{main_table_small}
\end{adjustbox}
\end{table}

\subsection{Ablation studies}\label{sec:ablation}

In the following, we discuss several experiments ablating aspects of our model as reported in Table~\ref{table:ablation}. 

\begin{table}
\centering

\caption{Relative performance across three ablation groups compared to the base model: (Top) Model size, language model, and other factors, using the base model with binning; (Middle) Clipping, context size, and fine-tuning, using the base model with one-dimensional embedding; (Bottom) Semantic analysis, using the base model with one-dimensional embedding. Improvements over the base model are highlighted in bold.}\vspace{1mm}
\label{table:ablation}
\begin{adjustbox}{max width=\textwidth}
\begin{tabular}{p{5.9cm}rrrrrrrrrrrrrr}
\toprule
\textbf{Experiment} & \multicolumn{1}{c}{\textbf{All}} &
\multicolumn{3}{c}{\textbf{CARTE}} &
\multicolumn{2}{c}{\textbf{OML-CC18}} &
\multicolumn{2}{c}{\textbf{OML-CTR23}} &
\multicolumn{3}{c}{\textbf{TabReD}} &
\multicolumn{3}{c}{\textbf{TALENT-Tiny}} \\
\cmidrule(lr){2-2} \cmidrule(lr){3-5} \cmidrule(lr){6-7} \cmidrule(lr){8-9} \cmidrule(lr){10-12} \cmidrule(lr){13-15}
&
\multicolumn{1}{c}{\textbf{Rank}} & \multicolumn{1}{c}{\textbf{Rank}} & \textbf{Acc} & \textbf{R$^\textrm{2}$} &
\multicolumn{1}{c}{\textbf{Rank}} & \multicolumn{1}{c}{\textbf{Acc}} &
\multicolumn{1}{c}{\textbf{Rank}} & \multicolumn{1}{c}{\textbf{R$^\textrm{2}$}} &
\multicolumn{1}{c}{\textbf{Rank}} & \multicolumn{1}{c}{\textbf{Acc}} & \multicolumn{1}{c}{\textbf{R$^\textrm{2}$}} &
\multicolumn{1}{c}{\textbf{Rank}} & \multicolumn{1}{c}{\textbf{Acc}} & \multicolumn{1}{c}{\textbf{R$^\textrm{2}$}} \\
\bottomrule\\
\makebox[3cm][l]{\textcolor{gray}{Model Size, Language Model, and others}} &&&&&&&&&&&&&&\\
\toprule
base model (binning) & 2.28 & 1.76 & 76.0 & 71.4 & 2.57 & 86.8 & 2.66 & 70.9 & 1.00 & 85.4 & 62.9 & 2.32 & 87.2 & 73.5\\
\midrule
large size & 2.76 & 2.18 & \bf{+0.2} & -0.2 & 4.01 & -5.3 & \bf{1.77} & -0.7 & 3.75 & -25.4 & -13.5 & \bf{1.84} & -0.1 & -0.0\\
medium size & 5.44 & 8.45 & -5.6 & -3.7 & 4.76 & -1.2 & 3.49 & -1.0 & 5.75 & -9.3 & -3.5 & 4.41 & -2.6 & -2.4\\
small size & 7.68 & 9.98 & -2.9 & -6.8 & 6.57 & -3.0 & 8.17 & -4.5 & 8.12 & -1.7 & -3.8 & 6.14 & -2.7 & -6.7\\
mini size & 8.10 & 10.69 & -4.2 & -8.4 & 7.15 & -4.0 & 8.06 & -4.2 & 5.50 & -0.3 & -3.5 & 7.00 & -3.7 & -6.9\\
\arrayrulecolor{mygrey}\midrule\arrayrulecolor{black}
multilingual-e5-small & 3.22 & 4.16 & -0.3 & -1.5 & 3.35 & -0.4 & 2.77 & -0.1 & 2.38 & -0.2 & -1.3 & \bf{2.27} & \bf{+0.7} & -0.7\\
gte-multilingual-base & 3.22 & 3.59 & \bf{+0.0} & -1.2 & 3.18 & -0.3 & 3.60 & -1.4 & 1.62 & -0.2 & -0.9 & 2.76 & \bf{+0.4} & -1.3\\
\arrayrulecolor{mygrey}\midrule\arrayrulecolor{black}
curriculum learning & 2.28 & \bf{1.51} & \bf{+1.1} & \bf{+0.1} & 3.10 & -0.2 & \bf{1.80} & -0.1 & 1.38 & \bf{+0.4} & -0.2 & 2.38 & -1.0 & \bf{+0.4}\\
\arrayrulecolor{mygrey}\midrule\arrayrulecolor{black}
clustering & \NAN & \NAN & -0.1 & \NAN & 3.82 & -1.5 & \NAN & \NAN & \NAN & -0.6 & \NAN & \NAN & -1.3 & \NAN\\
\arrayrulecolor{mygrey}\midrule\arrayrulecolor{black}
w/o bagging & 2.82 & 2.76 & -0.4 & -0.4 & 3.26 & -0.7 & 3.03 & -0.4 & 1.00 & \bf{+0.1} & -0.5 & \bf{2.24} & -0.6 & -0.0\\
\arrayrulecolor{mygrey}\midrule\arrayrulecolor{black}
ISAB & 5.48 & 6.16 & \bf{+0.3} & -3.1 & 5.57 & -2.1 & 5.51 & -0.9 & 3.88 & -0.2 & -2.7 & 4.70 & -0.5 & -6.1\\
\arrayrulecolor{mygrey}\midrule\arrayrulecolor{black}
non-shared weights & 4.36 & 5.61 & -0.1 & -2.6 & 4.29 & -0.9 & 3.09 & \bf{+0.8} & 5.88 & -1.0 & -2.6 & 3.65 & -0.7 & -1.2\\
\bottomrule\\
\makebox[3cm][l]{\textcolor{gray}{Clipping, Context Size, and Fine-tuning}} &&&&&&&&&&&&&&\\
\toprule
base model (1-dim) & 2.00 & 1.80 & 76.7 & 72.0 & 2.36 & 86.9 & 2.06 & 65.0 & 1.75 & 85.3 & 63.3 & 1.54 & 87.9 & 75.8\\
\midrule
0.1\% clipping & 2.03 & 1.80 & 0.0 & 0.0 & 2.36 & 0.0 & 2.37 & -0.3 & \bf{1.38} & 0.0 & \bf{+0.2} & 1.54 & 0.0 & 0.0\\
0.5\% clipping & 2.08 & \bf{1.76} & 0.0 & 0.0 & \bf{2.32} & -0.1 & 2.60 & \bf{+4.0} & \bf{1.38} & \bf{+0.1} & \bf{+0.1} & 1.70 & \bf{+0.1} & -0.2\\
2\% clipping & 2.97 & 2.88 & 0.0 & -0.5 & 2.43 & -0.1 & 5.03 & \bf{+5.8} & 1.75 & \bf{+0.1} & -0.5 & 2.43 & 0.0 & -2.0\\
binning & 4.04 & 4.69 & -0.7 & -0.6 & 3.21 & -0.1 & 6.34 & \bf{+5.9} & \bf{1.12} & \bf{+0.1} & -0.3 & 3.24 & -0.7 & -2.3\\
\arrayrulecolor{mygrey}\midrule\arrayrulecolor{black}
context=20000 & 2.06 & \bf{1.39} & \bf{+0.3} & \bf{+0.6} & 2.62 & -0.2 & \bf{1.66} & \bf{+4.2} & 3.25 & -0.4 & -0.5 & 2.00 & -0.4 & \bf{+0.5}\\
context=10000 & 2.00 & \bf{1.61} & \bf{+0.3} & \bf{+0.3} & 2.64 & -0.2 & \bf{1.57} & \bf{+4.2} & 1.75 & 0.0 & -0.1 & 1.76 & -0.2 & \bf{+0.5}\\
context=4096 & 2.81 & 3.55 & -0.1 & -0.6 & 3.03 & -0.3 & 2.11 & \bf{+4.0} & 2.25 & -0.3 & -0.1 & 2.16 & -0.4 & \bf{+0.4}\\
context=2048 & 4.11 & 6.41 & -1.0 & -1.9 & 3.61 & -0.4 & 3.06 & \bf{+3.6} & 2.50 & -0.1 & -0.7 & 3.24 & -0.9 & \bf{+0.1}\\
context=1024 & 5.49 & 9.10 & -1.3 & -3.8 & 4.33 & -0.5 & 4.60 & \bf{+2.7} & 2.62 & -0.4 & -1.6 & 4.24 & -1.3 & -0.6\\
\midrule
fine-tuning (per benchmark) & 2.60 & 2.51 & \bf{+0.3} & \bf{+0.1} & 3.1 & \bf{+0.2} & \bf{1.94} & \bf{+7.2} & 2.62 & \bf{+0.2} & -0.3 & 2.38 & -0.1 & \bf{+0.8}\\
\bottomrule\\
\makebox[3cm][l]{\textcolor{gray}{Semantics -- Features and Column Names}} &&&&&&&&&&&&&&\\
\toprule
base (feature and column name semantics)                              & 1.24 & 1.24 & 76.9 & 72.4 & \NAN & \NAN & \NAN & \NAN & \NAN & \NAN & \NAN & \NAN & \NAN & \NAN\\
\midrule
no feature semantics - Ordinal encoder         & 4.27 & 4.27 & -2.7 & -4.8 & \NAN & \NAN & \NAN & \NAN & \NAN & \NAN & \NAN & \NAN & \NAN & \NAN\\
no feature semantics - MinHash encoder        & 3.49 & 3.49 & -1.7 & -3.3 & \NAN & \NAN & \NAN & \NAN & \NAN & \NAN & \NAN & \NAN & \NAN & \NAN\\
no feature semantics - AutoGluon encoder      & 5.76 & 5.76 & -4.8 & -7.3 & \NAN & \NAN & \NAN & \NAN & \NAN & \NAN & \NAN & \NAN & \NAN & \NAN\\
no feature semantics - Gap encoder            & 4.12 & 4.12 & -1.4 & -4.3 & \NAN & \NAN & \NAN & \NAN & \NAN & \NAN & \NAN & \NAN & \NAN & \NAN\\
\midrule
column semantics  - add description     & \bf{1.20} & \bf{1.20} & 0.0 & 0.0 & \NAN & \NAN & \NAN & \NAN & \NAN & \NAN & \NAN & \NAN & \NAN & \NAN\\
column semantics  - drop column names & 3.35 & 3.35 & -1.2 & -2.1 & \NAN & \NAN & \NAN & \NAN & \NAN & \NAN & \NAN & \NAN & \NAN & \NAN\\
\bottomrule

\end{tabular}

\end{adjustbox}
\end{table}

\custompar{Semantics} Being our core contribution, we investigate the efficacy of the semantic cell and column header encoding used by ConTextTab. As we expect the results to be most pronounced in the presence of semantic features, we perform this ablation on CARTE exclusively.
To this end, we replace categorical and string features by ordinal encodings, as well as encodings via \texttt{MinHashEncoder} and \texttt{GapEncoder} (provided by the \texttt{skrub} library), and the AutoGluon feature encoder for comparison. As expected, we observe a significant drop in performance when discarding semantics completely or when using the conventional string encoders from the \texttt{skrub} library. Hence, ConTextTab successfully leverages the semantic content of features. We refer to Appendix~\ref{sec:further_ablations} for win ratio comparisons and p-value results to demonstrate statistical significance of our claim.

Secondly, we ablate the use of column header semantics in two ways: one, by replacing the column names with \texttt{col1}, ..., \texttt{colN}, effectively eliminating semantic content; as well as two, semantically enriching the column names by creating contextualized column descriptions using an LLM and replacing the column names with \texttt{<name>:<description>}. To this end, we used \texttt{gemma3-12b} and presented it with 5 randomly sampled rows, serialized as JSON, and prompting for contextualized descriptions of each column. 
Dropping column name semantics indeed results in a performance loss of about 1\% in accuracy and 2\% in $R^2$ score. This difference is significant with a win rate of 92\% of the base model and a vanishingly small p-value as shown in Appendix~\ref{sec:further_ablations}. Further enriching column header semantics slightly boosts performance, resulting in a slightly better rank, however with a win rate of 57\% at a p-value of 0.4 this is not statistically significant.

Overall, \ours{} successfully integrated and leverages semantics, both present in the table's features as well as those potentially contained in the table column names.

\custompar{Model size} As expected, decreasing from ``base'' to smaller models impacts performance. On the other hand, increasing to ``large'', thus doubling the number of layers and increasing the hidden dimension, has no significant impact on performance. As observed in Figure~\ref{fig:critical_difference_and_pretraining_size} (right), increasing the amount of training data leads to stagnating validation performance gains. We hypothesize that, in the current setup, our model is likely limited by the amount or diversity of data available with T4.

\custompar{Context size} The trained models can scale the input context size at inference. As expected, longer context improves results in a monotonic way, in exchange for longer computation time. In practice, the effects are stronger on larger datasets, simply because shorter context models (even with bagging) only get to see a subset of the training data. We stress again that this is entirely extrapolating outside of the pretraining setup, where the largest observed context has size 950. Nevertheless, with larger datasets and context, the model performance saturates, requiring further research in how to increase the capacity of tabular ICL.

\custompar{Training data} We ran training on randomly chosen subsets of the T4 dataset of $10^n$ tables, $n \in \{1, \dots, 6\}$. For this experiment we report the validation scores (on the validation held-out subset of the CARTE benchmark) in Figure \ref{fig:critical_difference_and_pretraining_size} (right). The results suggest that at least 100\,k of the tables are needed in order to train a model of state-of-the-art quality. While exact numbers may vary when using other datasets or model sizes, this gives a high-level estimate of the amount of required data, giving indication of why our model could not profit from a second step of curriculum learning on the much smaller dataset used by \citet{tabdpt}. Potentially, training is still limited by data amount or diversity, e.g.\ missing longer and wider tables, likely required to further scale tabular ICL.

\custompar{Classification target}
We compare the default cross-entropy loss for classification with the clustering approach described in Section~\ref{sec:alternatives}, allowing to retain the semantics of the target labels and lifting the limitation on the number of target classes as present in TabPFN or TabICL. However, most evaluation datasets only contain non-descriptive labels, and the number of classes is limited. Hence, the potential of clustering cannot emerge, and the default maintains a modest edge. We believe that semantically richer high-cardinality benchmarks are required to spur research in this direction.

\custompar{ISAB block} We observed comparable results on classification tasks and a modest drop in performance on regression tasks, while substantially enhancing efficiency by eliminating quadratic time complexity and reducing runtime by a factor of ten for large contexts.


\custompar{Bagging}
As expected, using bagging at inference time leads to consistent improvements. At the same time, in absolute terms, the gap is rather marginal, probably thanks to the fact that \ours{} gets rid of most sources of non-equivariance in the architecture (e.g.\ category indexing and positional encoding). As a result, depending on the use case, this allows to drop bagging, or reduce the number of bags, without compromising performance significantly. Given the great boost of performance when ensembling baselines via inner cross-validation, as reported in Appendix~\ref{app:additional-results} as well as recently observed in the literature~\cite{tabarena}, how to best leverage ensembled ICL models remains an open challenge.

\custompar{(No) weight sharing}
We observed a slight performance degradation when training the full model as opposed to using weight sharing, which may be due to either an insufficient training duration, suboptimal selection of training hyperparameters, or limited pretraining dataset size and variety.

\custompar{Further ablations} Other changes (the used sentence embedding model, longer context in training, target clipping for regression, as well as benchmark-dependent fine-tuning) did not result in significant performance gains. Therefore, we refer to the Appendix \ref{sec:further_ablations} for further discussion of those results.




\section{Conclusions}
\custompar{Limitations and future work}
While achieving SOTA results across the investigated datasets, we observe several limitations of our proposed approach. 
One drawback of using real-world data for training is the possibility of contamination, e.g.\ the presence of evaluation tasks in the training corpus. Since the CARTE benchmark is our focus, we conducted a contamination study, using the column name and cell value embeddings created by a text embedder and matching similarities. We did not find any contamination of CARTE in T4. For contamination of OpenML datasets, we refer to the original study by~\citep{tabula8b}. Either way, as a single table is only seen a few times during training, we believe that memorization is likely not a practical problem when training on real-world data.

Generally, all investigated table-native ICL models, namely ours, TabPFN, TabDPT, and TabICL, fail to scale their performance to very large datasets. Increasing the context length as well as bagging did not fully resolve these issues. Using local context, as done in TabDPT, might overcome such limitations but results are likely limited by the row-based architecture in that case. In the large-data case, conventional methods, in particular when stacked and ensembled via AutoGluon, still perform best. Overcoming this remains one of the major research challenges for tabular foundation models.

We observe little gain when training on larger data corpora, e.g.\ increasing the pretraining subset of T4 or adding tables from the AutoML benchmark as used for pretraining TabDPT. Together with the investigated scaling of our model, we believe that more diverse real-world data is needed to fully unlock semantic understanding at scale. While models trained on synthetic data do not face this issue in principle, how to incorporate semantic alignment in these models remains an open question. Similarly for evaluation, semantically rich benchmarks are scarce. While CARTE includes rich semantic features, all classification targets are binary, containing no semantic information. Furthermore, it contains mostly small to medium sized datasets, in particular in terms of the number of features. Hence, we argue that more diverse, semantically rich data is required -- both at scale for pretraining larger models with longer context, as well as curated for evaluation, where recent works show promising first steps in this direction~\citep{texttabbench}.

Finally, our investigation to utilize target semantics and lift the strict class limit at inference prevalent in current tabular ICL, by using a dynamic supervised clustering approach, did not result in performance gains. This might relate to the lack of semantically meaningful targets in current tabular benchmarks as well as the limited target cardinality across them and thus remains an open problem.




\custompar{Summary} We have introduced \ours{} -- a context aware table-native in-context learner trained on real-world data for both classification and regression tasks. Evaluated across a wide range of tabular benchmarks, our model performs competitively on a range of non-semantic benchmarks while achieving state-of-the-art results, significantly outperforming existing table-native ICL approaches such as TabPFN, on the semantically rich CARTE benchmark and low-data regime. 

\section*{Acknowledgements}
We would like to thank Johannes Hoehne, Johannes Hoffart, and Markus Kohler for their insightful comments and suggestions throughout the development of this work, which have greatly contributed to shaping its direction and quality. Furthermore, we thank Mahdi Hadj Ali for creating the semantically enriched evaluation datasets used in our semantic ablation as well as Annie Lim for running the ablation. Finally, we thank Frank Essenberger and G\"unther Schindler for their support with our data and compute infrastructure.



{
\small

\bibliography{refs}
\bibliographystyle{acl_natbib}

}

\newpage

\appendix

\section{Further Results}\label{app:additional-results}

\subsection{Relation between dataset size and model performance}\label{app:additional-results-data-size}

\begin{figure}[H]
    \centering
    \includegraphics[width=0.98\linewidth]{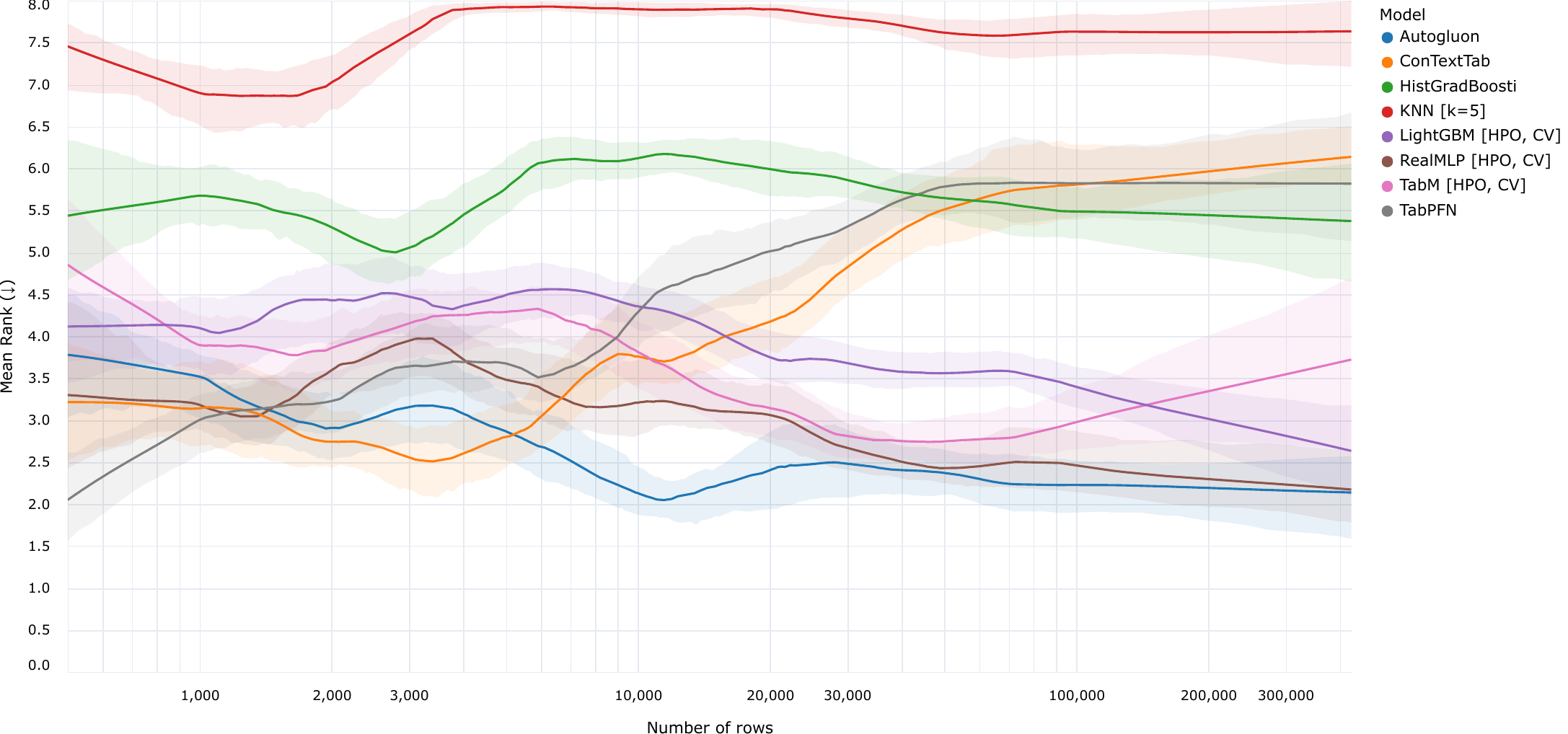}
    \caption{Relation between number of training dataset rows and performance, obtained as a LOWESS regression in the plane $\log(n_{\mathrm{rows}}, \textnormal{rank})$. The confidence bands are the 80\% confidence intervals obtained via bootstrapping.}
    \label{fig:rank_vs_num_rows}
\end{figure}

We plot the average rank of each model as a function of the dataset size (expressed in number of rows) across all evaluated datasets in Figure~\ref{fig:rank_vs_num_rows}. Note that TabICL is missing from this comparison as it only handles classification tasks. We can observe that, as expected, \ours{} and TabPFN excel in the low data regime, while AutoGluon takes the lead when enough data is available. In the very low data regime below 1000 training rows, TabPFN performs best, however our model surpasses its performance for larger datasets with more than 1000 rows. Overall, \ours{} remains competitive with AutoGluon until roughly 10\,000 rows, possibly as a result of the limitation in both the inference context size and the context size seen during training. After 10\,000 rows, gradient boosting methods, as well as RealMLP, start surpassing the performance of \ours{} as well as TabPFN. Overall, this indicates the need of further research for tabular ICL to handle larger amounts of training data, but also the availability of more diverse benchmarks, covering larger datasets, as the current evaluation is dominated by datasets with less than 10\,k rows and less than 100 columns, see Figure~\ref{fig:dataset-statistics}.

\subsection{Extended results}\label{app:additional-results-extended}

\custompar{Extended baselines}
Including further baselines not depicted in the main paper, Table~\ref{table:ablation_all_results} shows all collected evaluations across the investigated benchmarks. In addition to the models shown in the main section, here we also depict performance of (tuned) default variants of XGBoost, LightGBM, CatBoost, RealMLP, and TabM. We also report results for Mambular~\cite{mambular}, however solely on CARTE and TALENT-Tiny as evaluation took too long on larger datasets due to slow convergence.

\begin{figure}
    \centering
    \begin{subfigure}[b]{\textwidth}
        \includegraphics[width=0.48\textwidth]{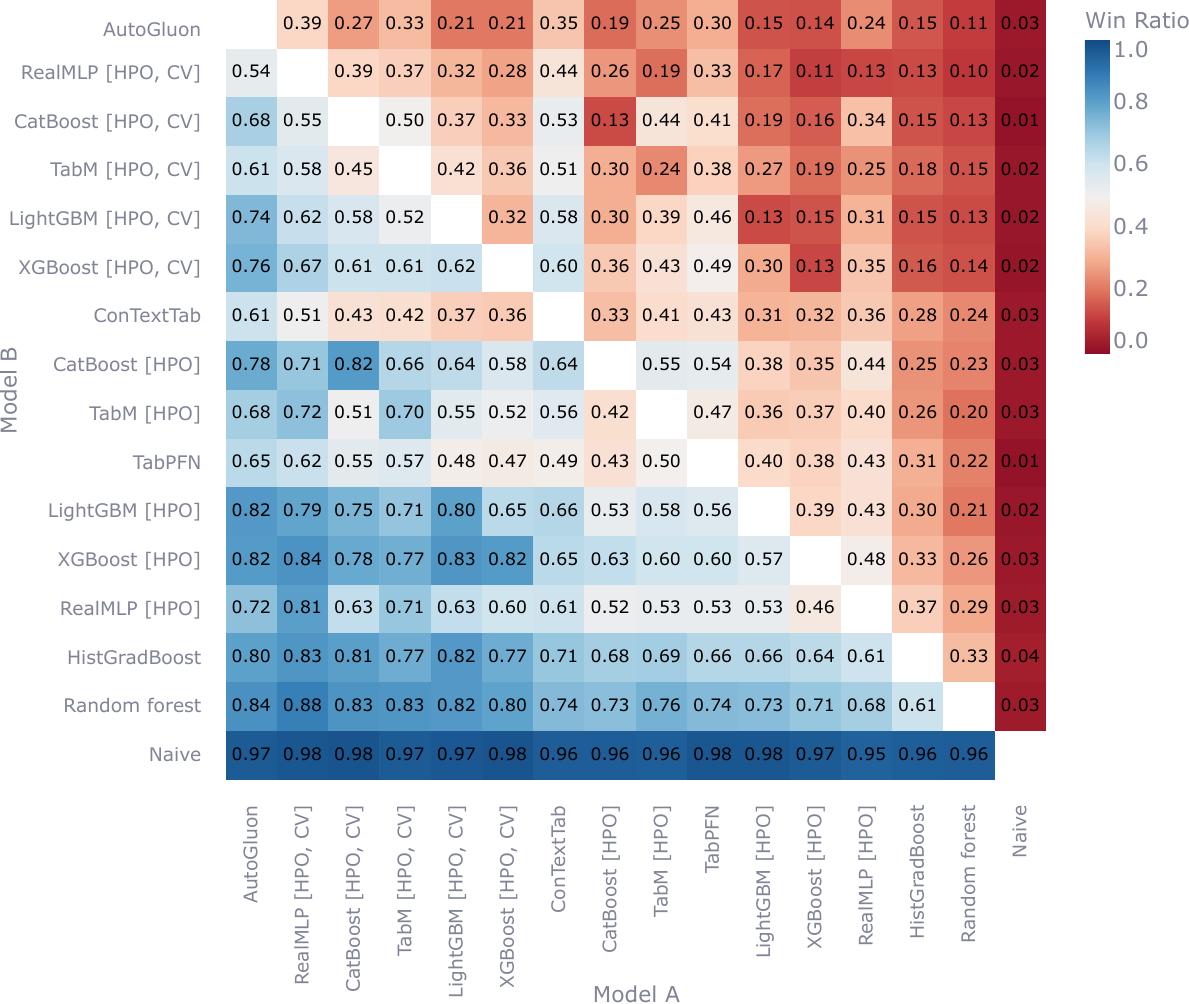}\hfill
        \includegraphics[width=0.48\textwidth]{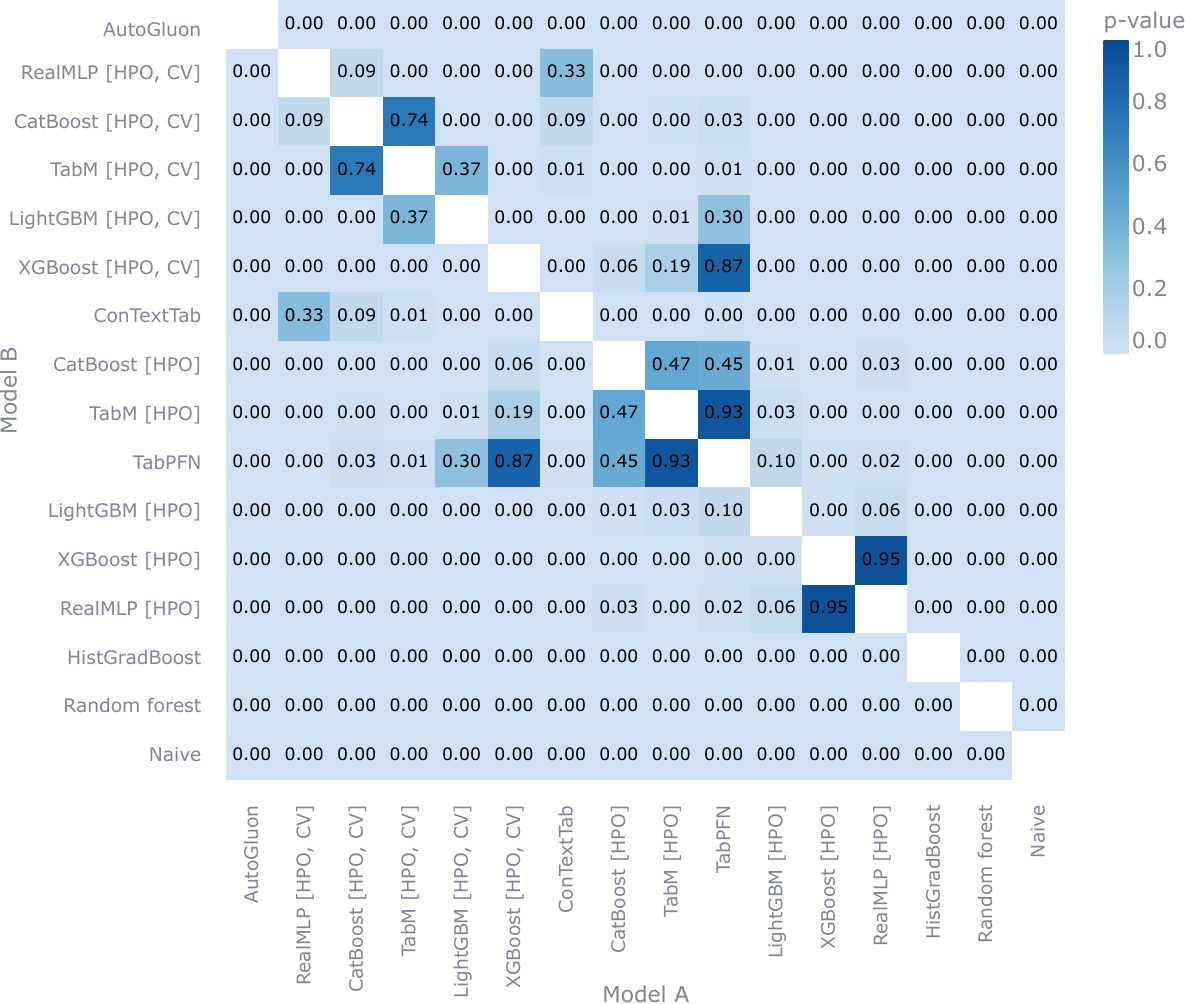}
        \caption{Win ratio matrix with Model A wins over Model B (left) and p-values (right).}
    \end{subfigure}\\[1mm]
    \begin{subfigure}[b]{0.7\textwidth}
        \includegraphics[width=\textwidth]{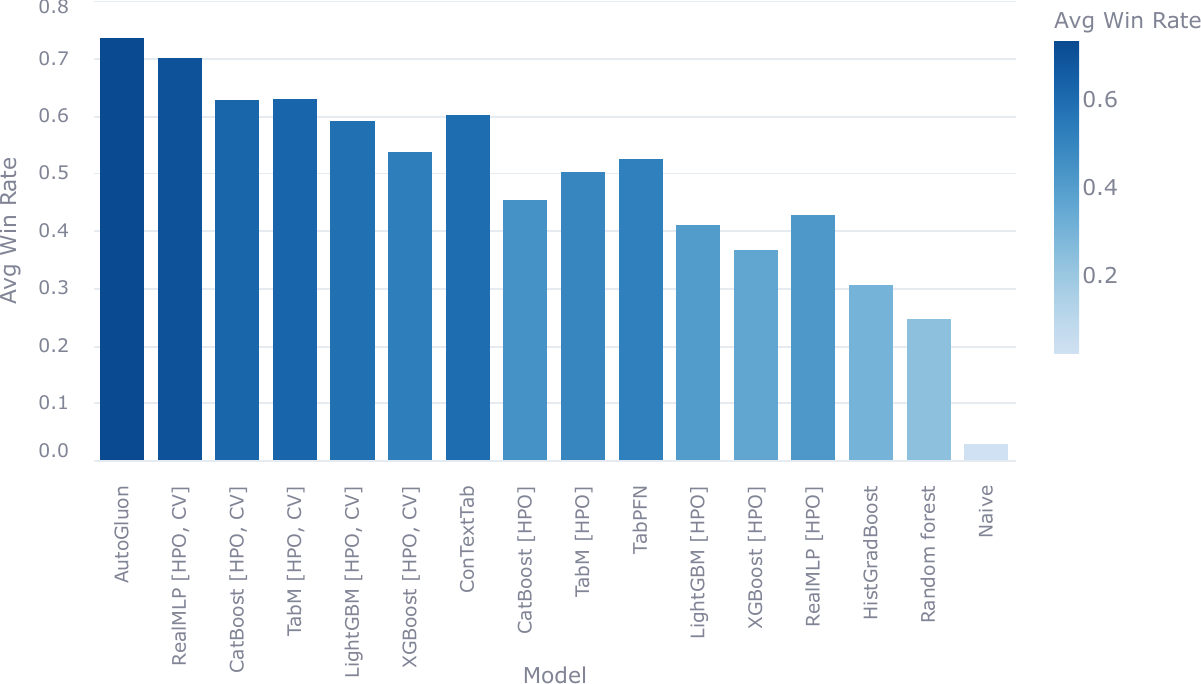}
        \caption{Average Win ratio of Model A against all others.}
    \end{subfigure}
\caption{Win ratio confusion matrix and average of the investigated models across all 203 datasets.  Wins are calculated based on accuracy on classification and $R^2$ on regression datasets. Ties are not counted as wins. Models are sorted by descending overall rank.}
\label{fig:win-ratios}
\end{figure}

\custompar{Win ratios, significance scores and CD diagrams}
Additionally, we show the win ratio confusion matrix, p-values, and averages of the investigated models across all investigated benchmarks in Figure~\ref{fig:win-ratios}.

\begin{figure}
    \centering
    \begin{subfigure}[b]{0.48\textwidth}
        \includegraphics[width=\textwidth,clip,trim={0 0 0 8mm}]{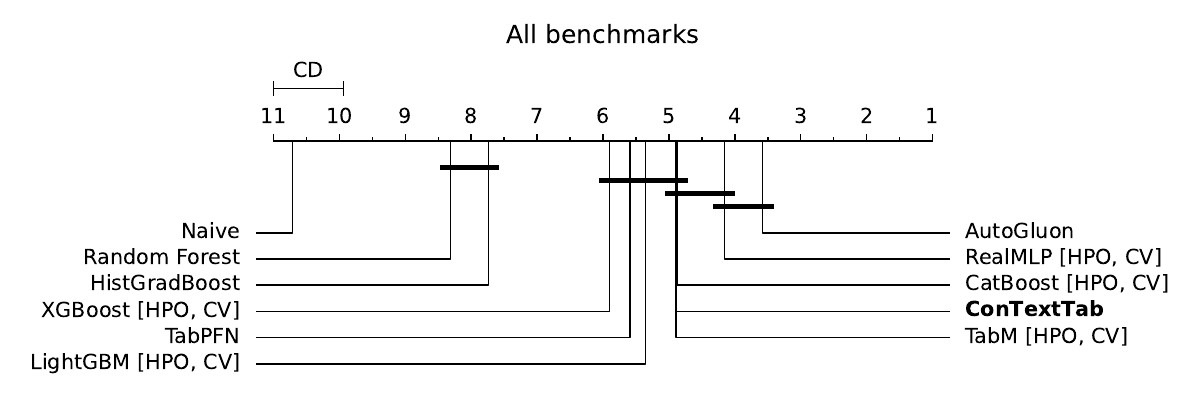}
        \caption{All benchmarks.}
    \end{subfigure}
    \hfill
    \begin{subfigure}[b]{0.48\textwidth}
        \includegraphics[width=\textwidth,clip,trim={0 0 0 8mm}]{figures/cd_diagrams/REBUTTAL_cd_diagram_carte.pdf}
        \caption{CARTE.}
    \end{subfigure}\\
    \begin{subfigure}[b]{0.48\textwidth}
        \includegraphics[width=\textwidth,clip,trim={0 0 0 8mm}]{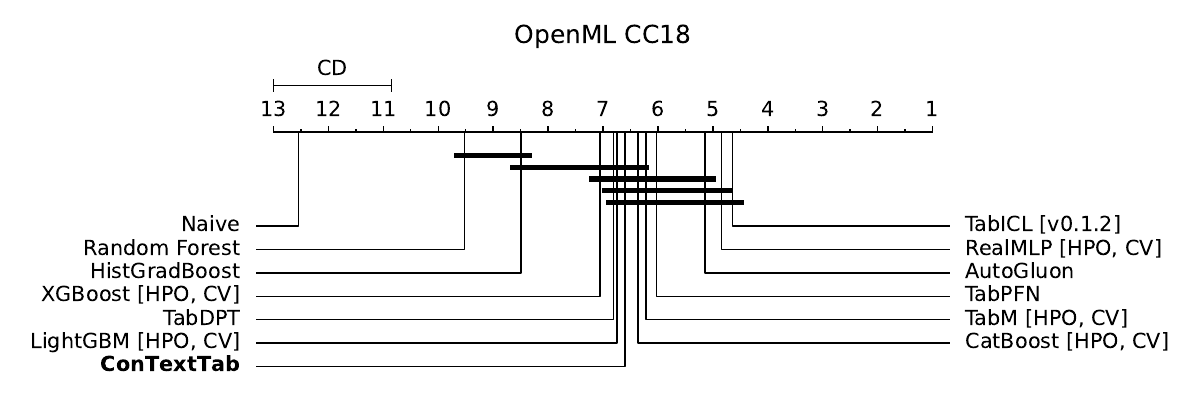}
        \caption{OpenML-CC18.}
    \end{subfigure}
    \hfill
    \begin{subfigure}[b]{0.48\textwidth}
        \includegraphics[width=\textwidth,clip,trim={0 0 0 8mm}]{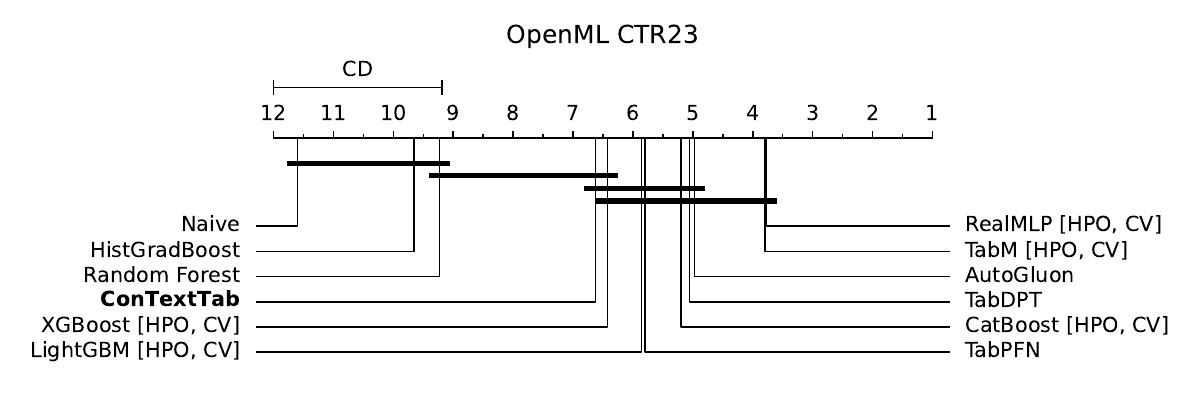}
        \caption{OpenML-CTR23.}
    \end{subfigure}\\
    \begin{subfigure}[b]{0.48\textwidth}
        \includegraphics[width=\textwidth]{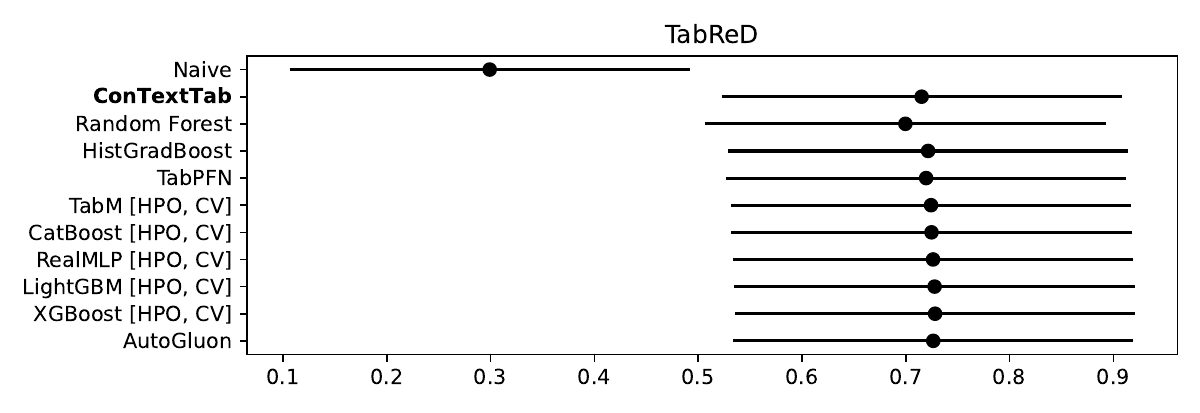}
        \caption{TabReD.}
    \end{subfigure}
    \hfill
    \begin{subfigure}[b]{0.48\textwidth}
        \includegraphics[width=\textwidth,clip,trim={0 0 0 8mm}]{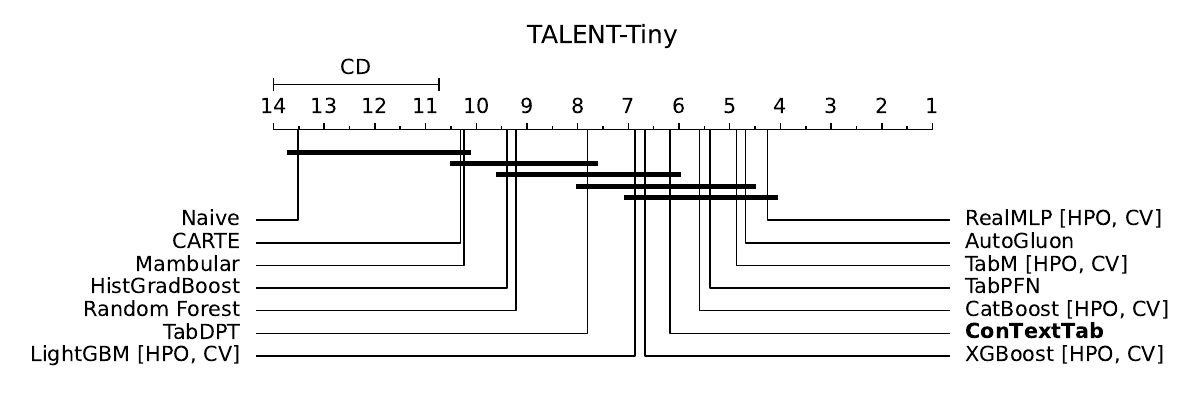}
        \caption{TALENT-Tiny.}
    \end{subfigure}
\caption{Critical difference diagrams for all evaluated benchmarks collectively and for each benchmark individually. As TabReD only contains 8 datasets, only ranking results are depicted.}
\label{fig:cd-diagrams-all-benchmarks}
\end{figure}

For fine-grained insights, we also plot per-benchmark results in Figure~\ref{fig:win-ratio-all-benchmarks} for win ratios and Figure~\ref{fig:cd-diagrams-all-benchmarks} for CD diagrams.

Across all benchmarks investigated, ConTextTab shows very strong performance, outperforming other ICL such as TabPFN and TabICL as well as tuned trees as well as RealMLP and TabM. However, hyperparameter optimized baselines with inner CV ensembles show very strong performance on non-semantic benchmarks (OpenML CC18/CTR23, TabReD, and TALENT-Tiny) where they outperform ConTextTab. This, of course, comes with much longer training times. Furthermore, detailed 1 vs 1 investigation, as well as the CD diagrams, paint a more nuanced picture, showing that differences in performance to \ours{} are often not statistically significant.

On the other hand, detailed evaluation on the semantically rich CARTE benchmark show that \ours{} significantly outperforms all other models and even performs compatible to AutoGluon. With a win ratio of 65\% and p-value of 0.03 over \ours{}, this gain is not as significant as it might appear from the average scores reported in Table~\ref{table:main_results} or Table~\ref{tab:all-dataset-tables}.
Notably, \ours{} outperforms all tabular ICL methods in the presence of semantically rich features, for example TabPFN: with a win rate of 96\% of \ours{} and a vanishingly small p-value, this is statistically significant. On non-semantic datasets, TabPFN wins over \ours{} with rates between 49\% and 66\%, however, with p-values between 0.09 and 0.64, the differences are not statistically significant.

\begin{figure}
    \centering
    \begin{subfigure}[b]{\textwidth}
        \includegraphics[width=0.475\textwidth]{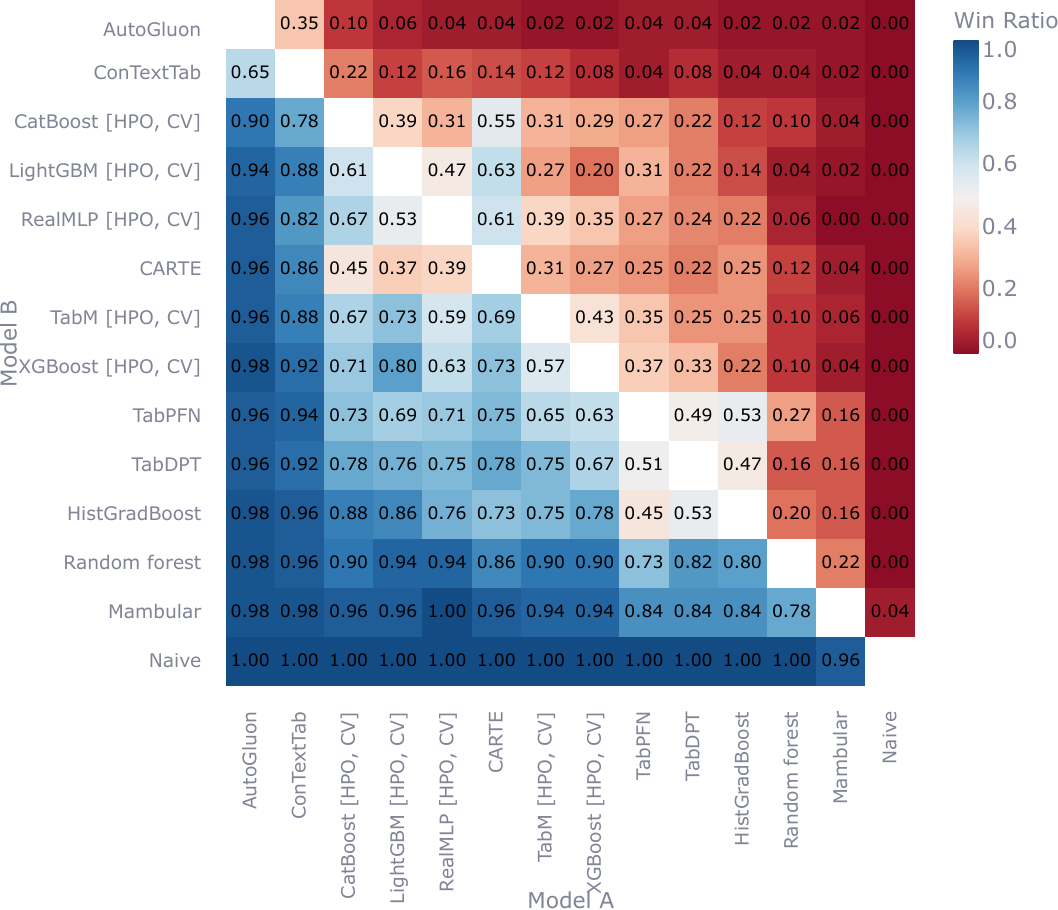}\hfill
        \includegraphics[width=0.475\textwidth]{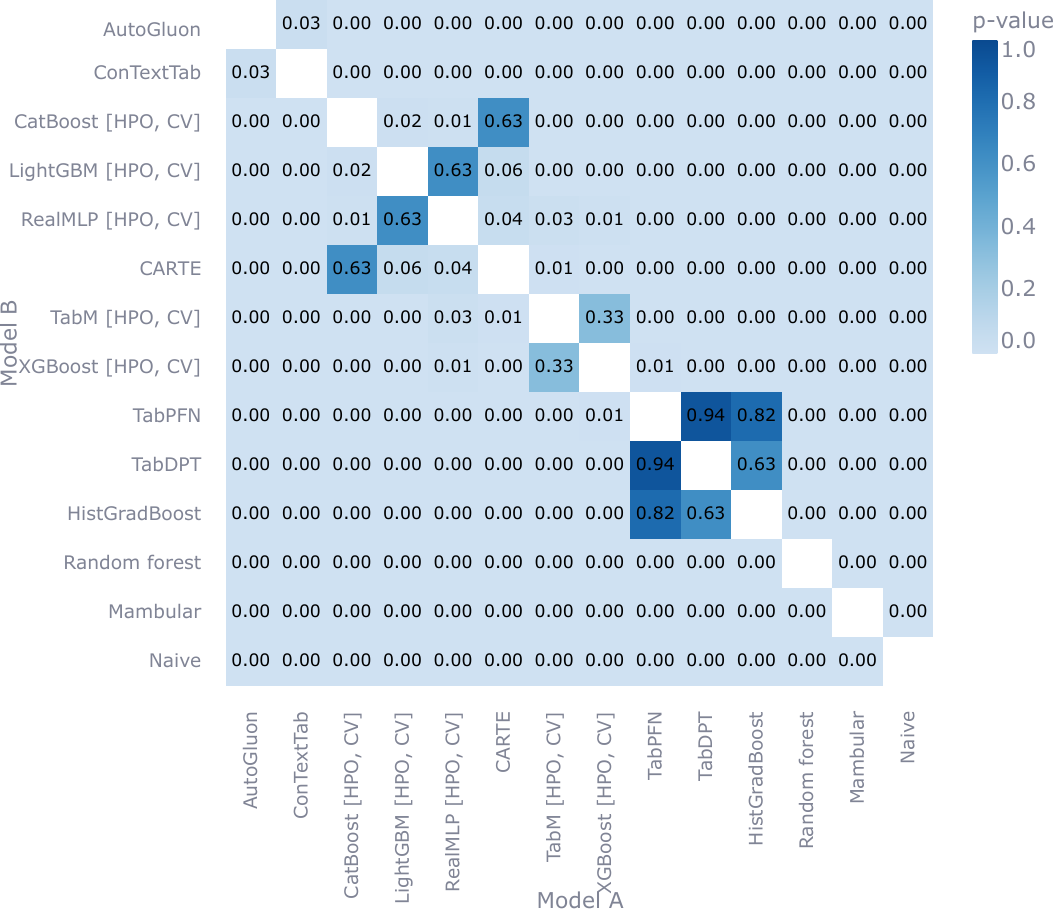}
        \caption{CARTE.}
    \end{subfigure}\\[5mm]
    \begin{subfigure}[b]{\textwidth}
        \includegraphics[width=0.475\textwidth]{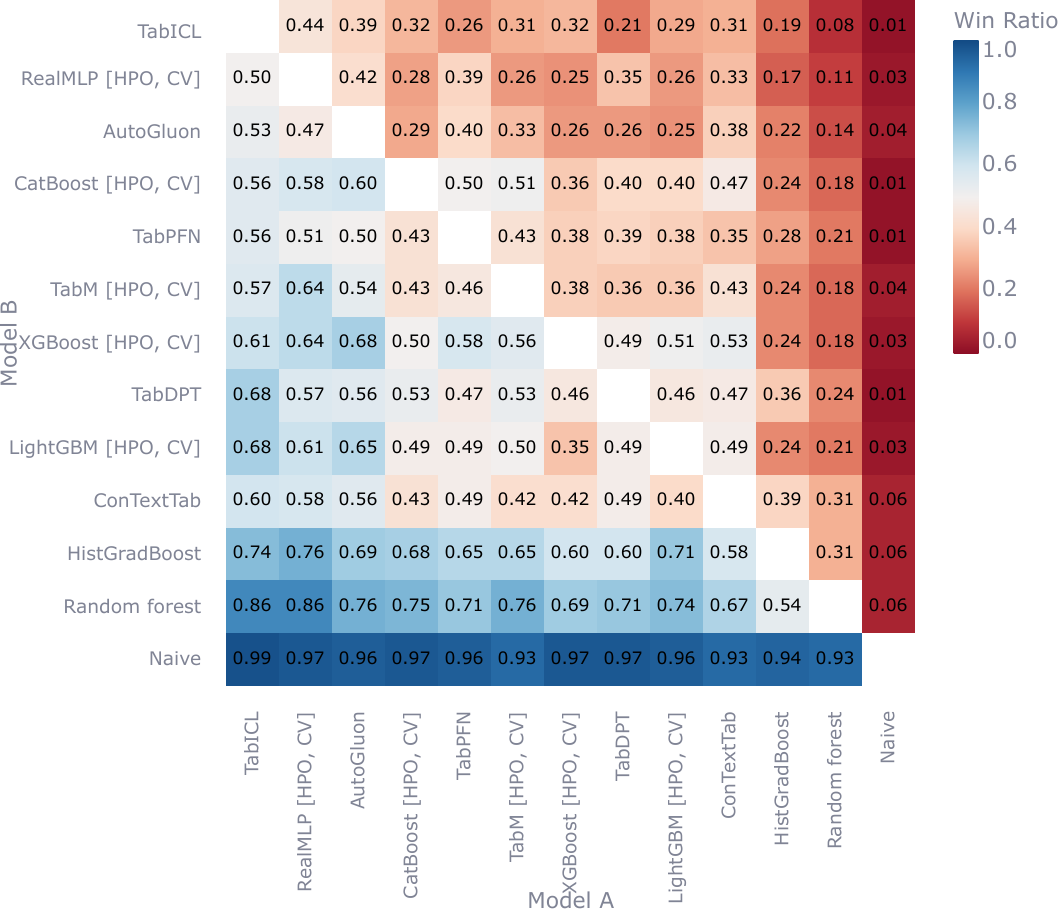}\hfill
        \includegraphics[width=0.475\textwidth]{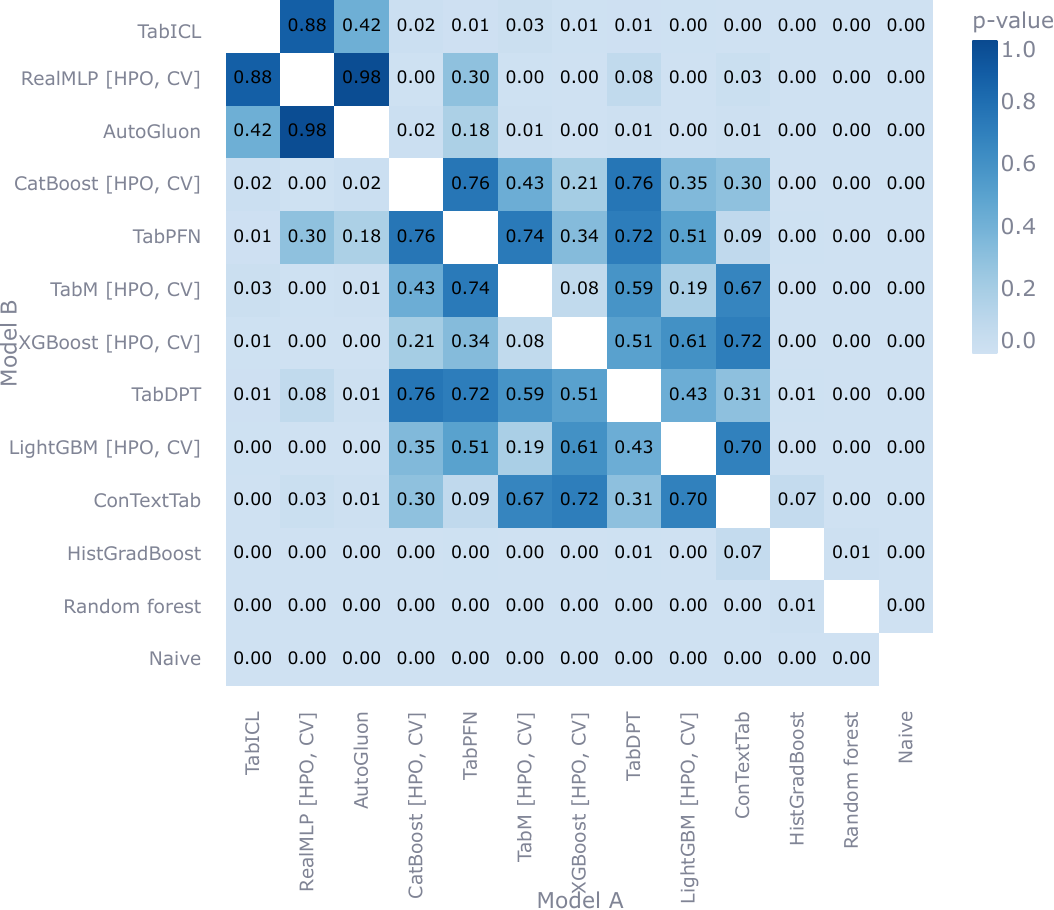}
        \caption{OpenML-CC18.}
    \end{subfigure}\\[5mm]
    \begin{subfigure}[b]{\textwidth}
        \includegraphics[width=0.475\textwidth]{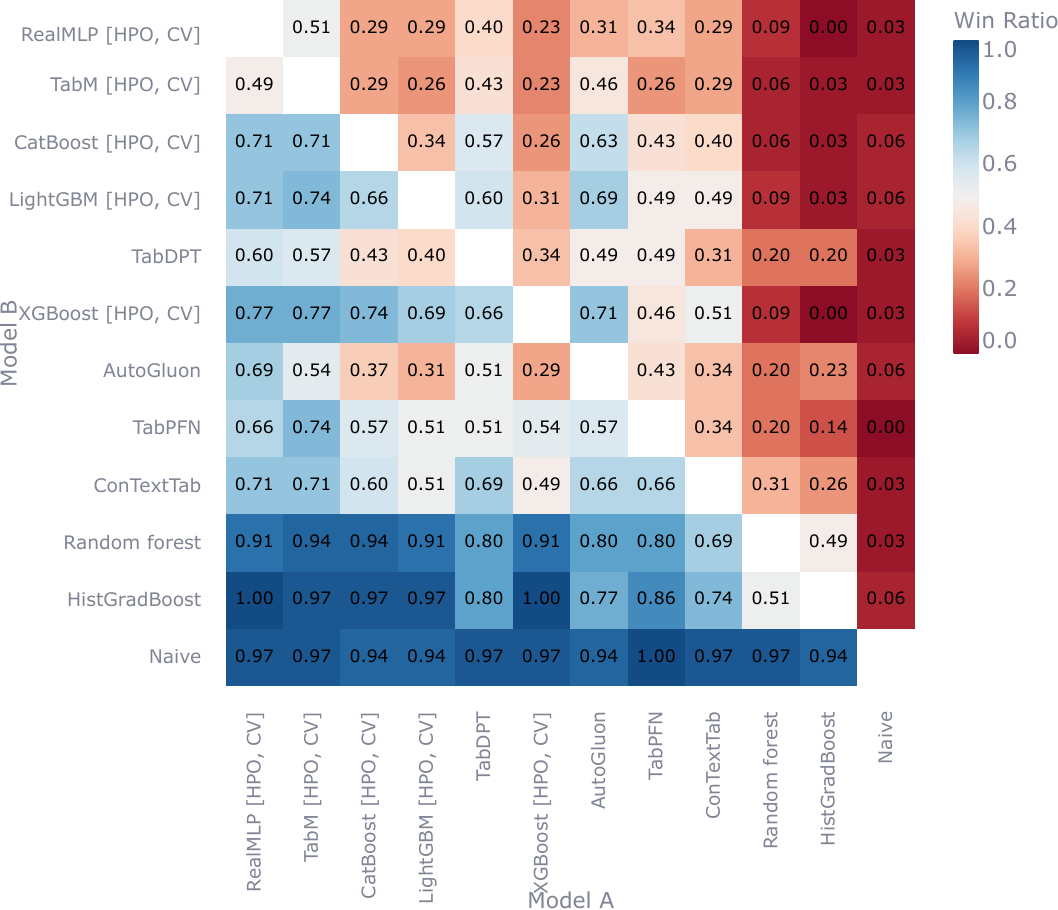}\hfill
        \includegraphics[width=0.475\textwidth]{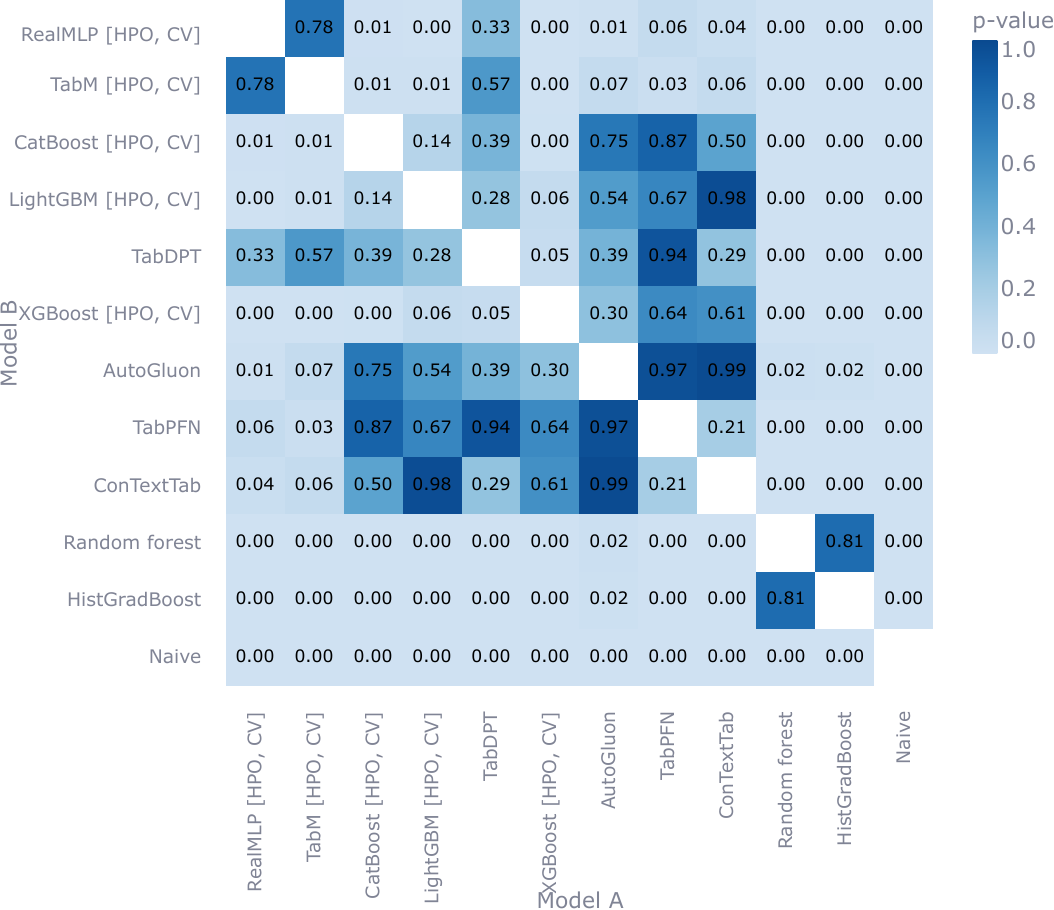}
        \caption{OpenML-CTR23.}
    \end{subfigure}
\end{figure}
\begin{figure}\ContinuedFloat
    \centering
    \begin{subfigure}[b]{\textwidth}
        \includegraphics[width=0.475\textwidth]{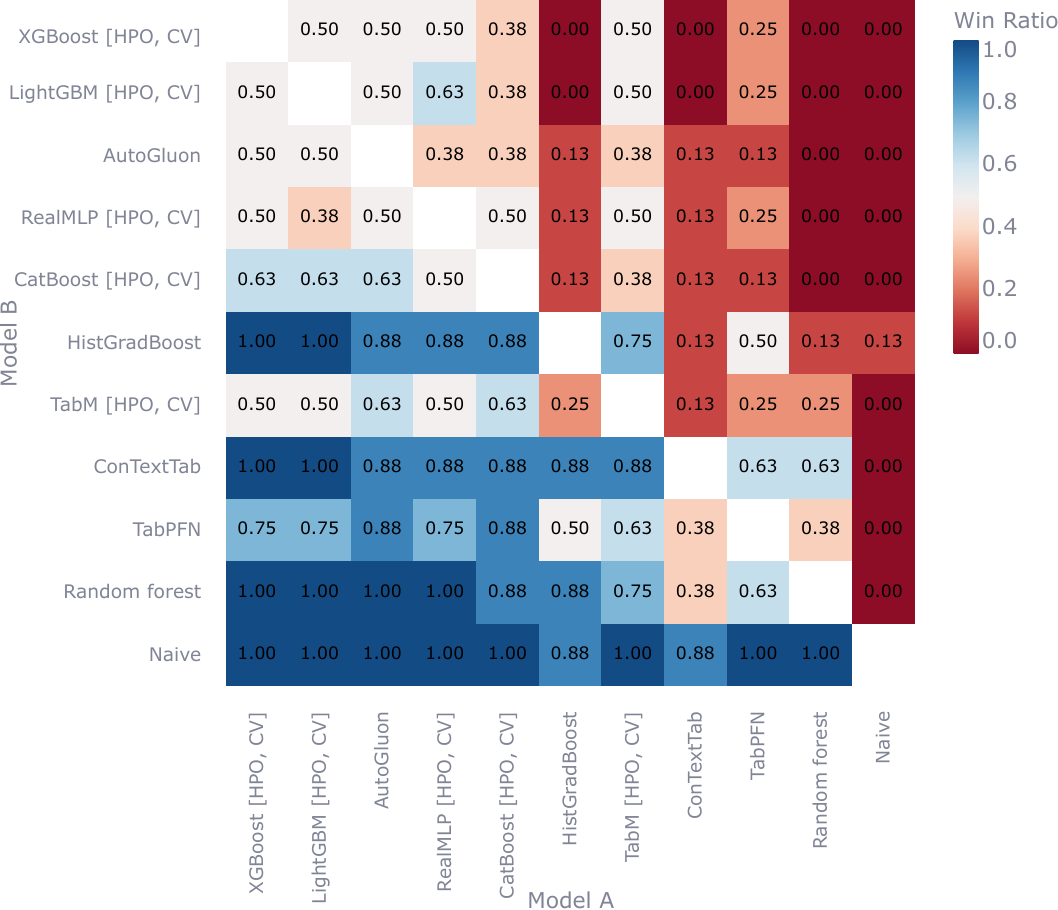}\hfill
        \includegraphics[width=0.475\textwidth]{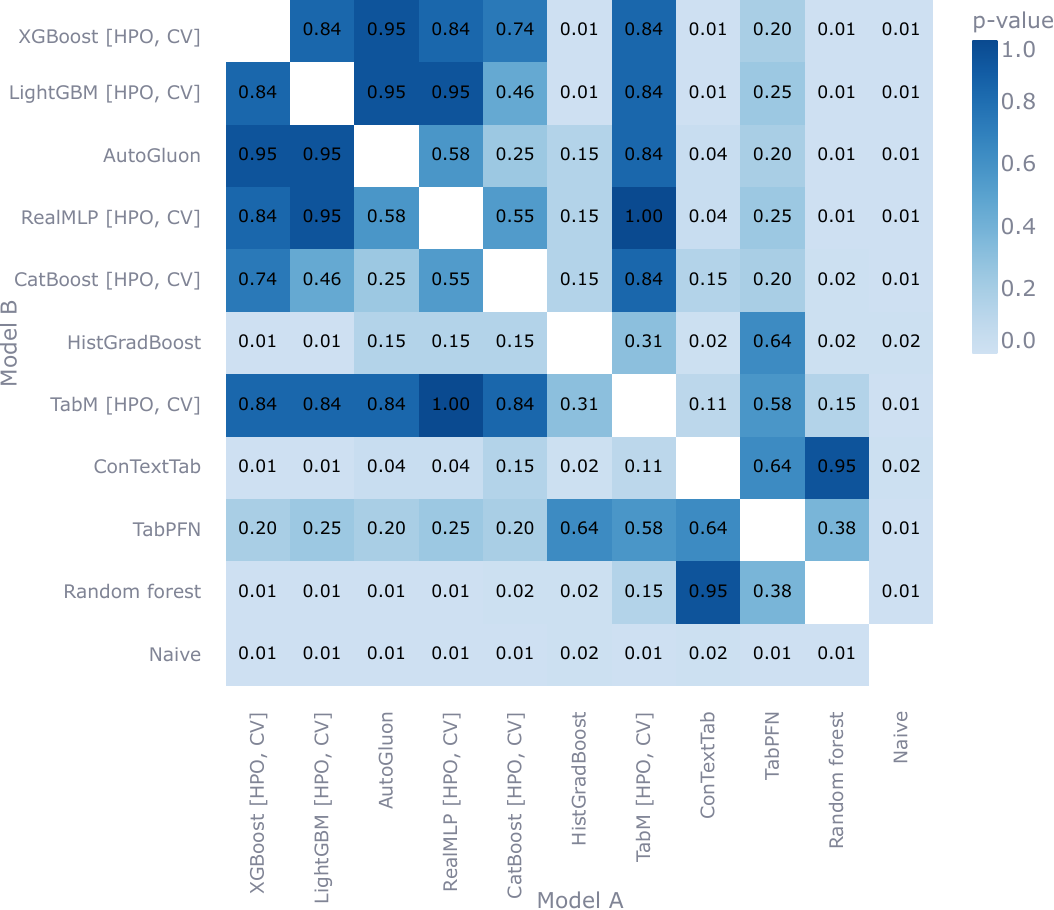}
        \caption{TabReD.}
    \end{subfigure}\\[5mm]
    \begin{subfigure}[b]{\textwidth}
        \includegraphics[width=0.475\textwidth]{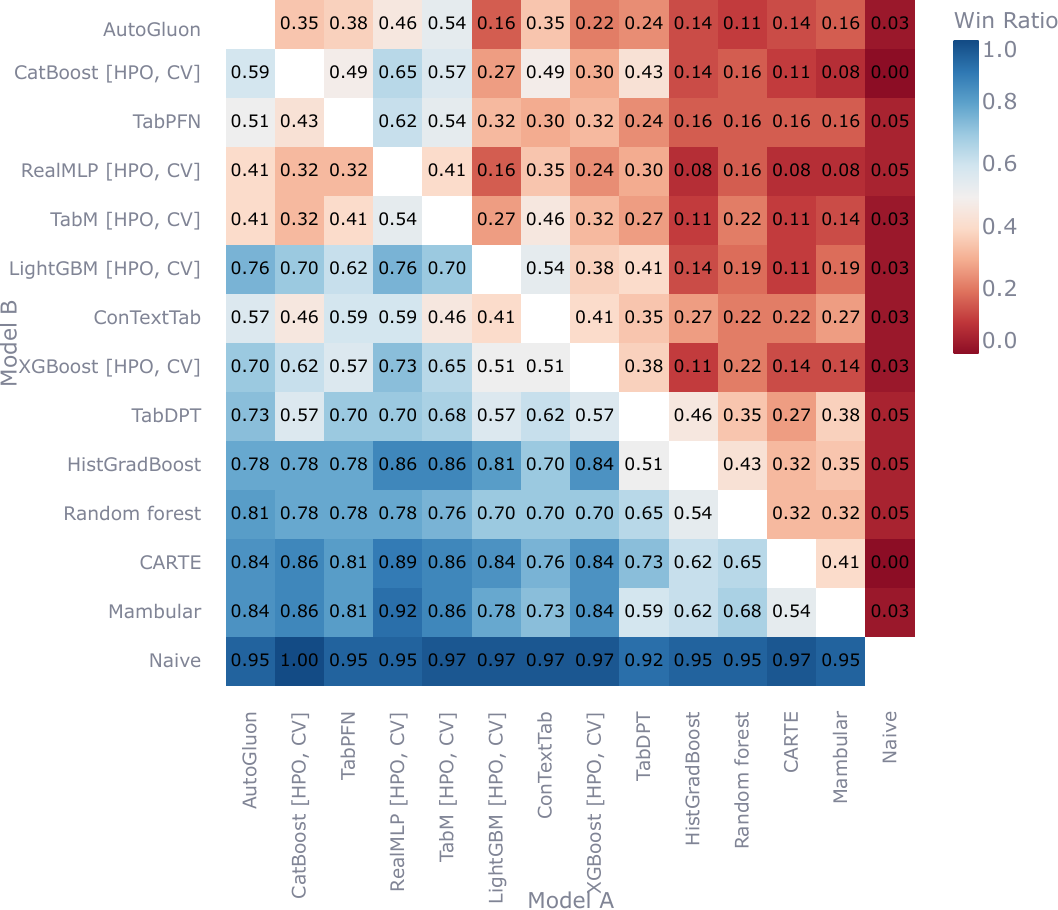}\hfill
        \includegraphics[width=0.475\textwidth]{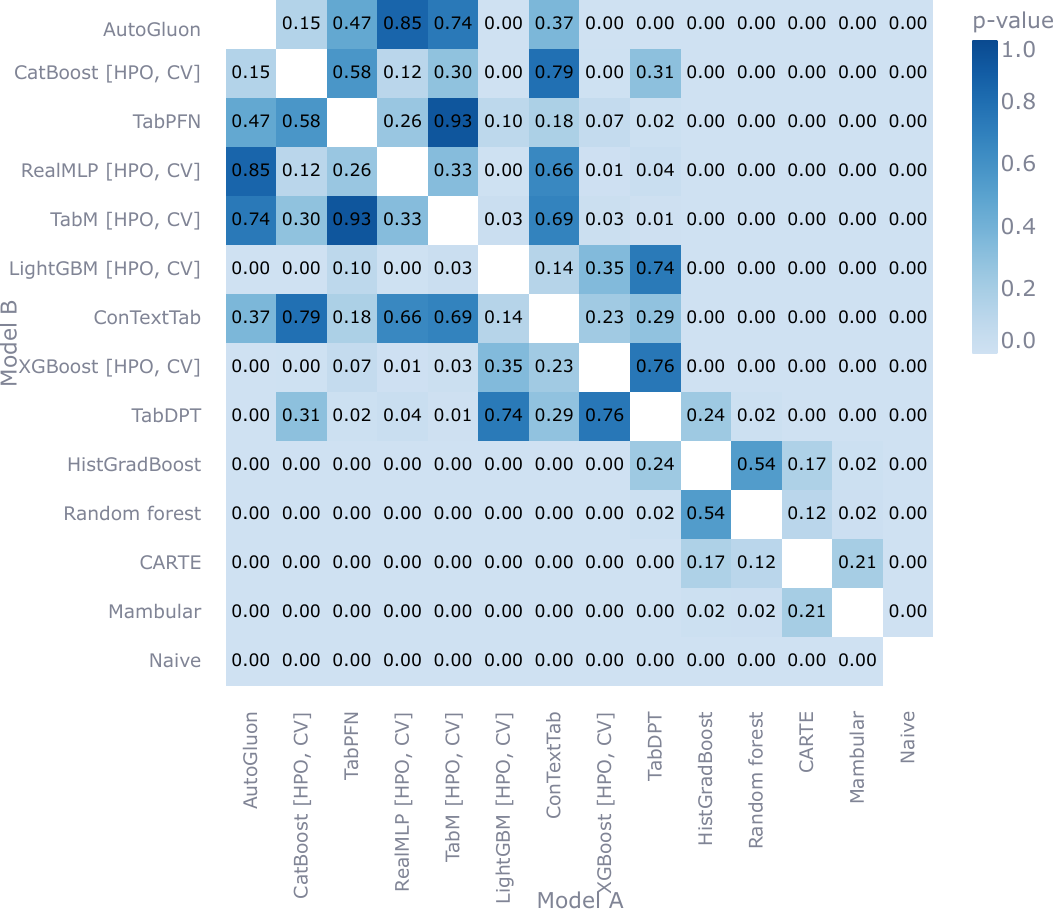}
        \caption{TALENT-Tiny.}
    \end{subfigure}
\caption{Win ratios (Model A wins over model B) as well as p-values of the two-sided Wilcoxon signed-rank test across evaluated baselines separately for each benchmark. Wins are calculated based on accuracy on classification and R2 on regression datasets. Ties are
not counted as wins. Models are sorted by descending overall rank.}
\label{fig:win-ratio-all-benchmarks}
\end{figure}

\setlength{\arrayrulewidth}{1.2pt}  
\renewcommand{\arraystretch}{1.1}
\begin{table}
\caption{Performance comparison across all evaluated benchmarks, depicting mean accuracy (Acc) for classification and (soft-clipped) $R^2$ score for regression tasks, in percent. Missing values, due to architectural limitations or failed evaluations, are denoted as N/A and excluded from the rank calculations. Models are sorted according to their ranking on the CARTE benchmark.}\vspace{1mm}
\label{table:ablation_all_results}
\centering
\begin{adjustbox}{max width=\textwidth}
\input{main_table}
\end{adjustbox}
\end{table}

\clearpage
\subsection{Runtime analysis}\label{app:runtime-analysis}
\begin{figure}[H]
    \centering
    \includegraphics[width=0.98\linewidth]{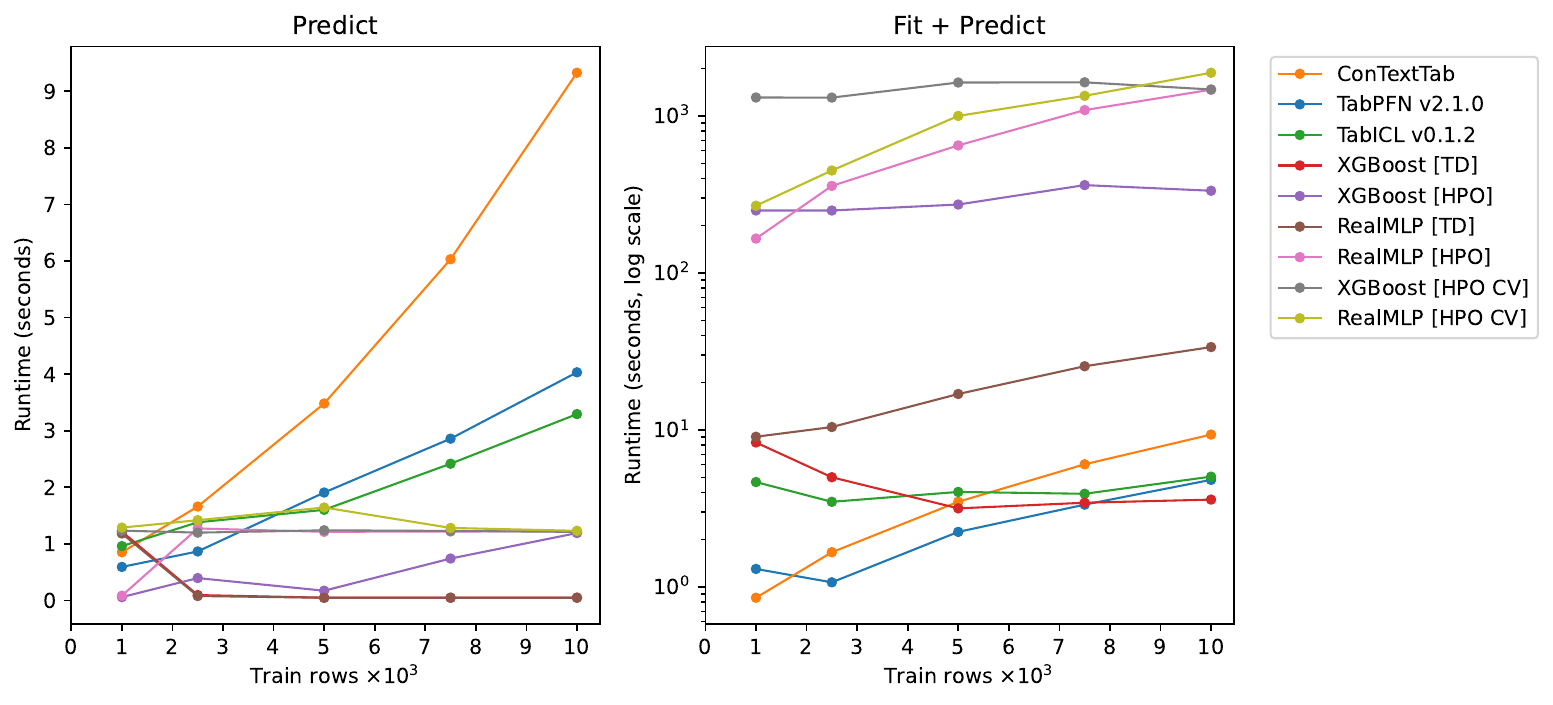}
    \caption{Runtime comparison of different models across varying numbers of training rows. (Left) Runtime for model prediction. (Right) Runtime for model prediction, including fitting. Note the linear y-axis scale on the left side as opposed to the log-scale on the right.}
    \label{fig:timing_analysis}
\end{figure}

We performed a runtime analysis of \ours{}, comparing it against TabPFN, TabICL, and two non-ICL models, XGBoost and RealMLP. The results are presented in Figure~\ref{fig:timing_analysis}. For this analysis, we created a synthetic dataset consisting of 100 columns, with an equal number of categorical/textual and numerical features, and varied the number of training/context rows from 1,000 to 10,000. We conducted tests by running one test/query row at a time with varying numbers of training/context rows, using 5 repeats and reporting averages for each table size. The experiments were conducted on a compute node with 40 CPU cores, 320\,GB of RAM and an H100 GPU with 96\,GB of VRAM.

We find that the throughput of \ours{} is comparable to other tabular ICL models, such as TabPFN and TabICL, with runtimes ranging from 0.1s to 9s, depending on the table size. As expected, due to the LLM embedding overhead, \ours{} is typically slightly slower than TabPFN or TabICL. For ConTextTab, we observe quadratic scaling behavior, as expected from its transformer backbone with memory-efficient attention. However, for TabPFN and TabICL, this effect is not noticeable because of the use of flash attention. As a result, ConTextTab is roughly twice as slow for larger contexts with \num{10000} rows. Moreover, inference runtime can be improved with additional technical approaches, such as KV caching or model compilation. 

Note that, while non-ICL models offer significantly faster prediction times, they require extensive training. In particular, the hyperparameter-optimized (HPO) and CV-ensembled (HPO,CV) versions of XGBoost and RealMLP exhibit fit+predict times that are two orders of magnitude longer.

\subsection{Further ablations} \label{sec:further_ablations}
In the following, we discuss in more detail the ablations as presented in Table~\ref{table:ablation}.

\custompar{Text embedding model}
We test the performance of two more recently released sentence embedding models, namely \texttt{Multilingual e5 small}~\citep{e5} and \texttt{gte-multilingual-base}~\citep{mgte}. However, we cannot observe significant impact on the final results over the default \texttt{all-MiniLM-L6-v2} sentence embedder. This potentially indicates the necessity of larger, semantically rich training data or further research into aligning these sentence embedders with a tabular foundation model. 

\custompar{Curriculum training with longer context}
In this setup, after training the model on T4 with a context size of up to 950, training is extended by including also the dataset used by \citet{tabdpt}, collected from the OpenML AutoML benchmark, and extending the context size up to 4000 rows. Data is sampled from either T4 with 80\% probability or the dataset from by \citet{tabdpt} with 20\% probability. The resulting model shows some improvements on a few tasks, but by statistically insignificant amounts.

\custompar{Regression target and clipping}
When using a one-dimensional encoding for numerical values, we can choose to clip the tails of the distribution for better stability. This is needed in pretraining due to the noisy data. In practice, whether to do this on downstream tasks can be chosen depending on outlier analysis of the data itself. We test alternative levels of clipping: no clipping (default), clipping at 0.1\%, 0.5\% and 2\% quantiles. We also compare this to using binning instead of one-dimensional embeddings. We find that both clipping by 2\% and binning cause a slight reduction in quality (since we are using $n=16$ bins, note that binning also incurs in clipping at $\frac{1}{32} \approx 3\%$). The only exception is in the OpenML-CTR23 benchmark: it contains one dataset with extreme outliers where no model achieves a positive $R^2$ score, and adding clipping there limits degenerate scores.

\custompar{Impact of fine-tuning}
The training process can be easily adapted to fine-tuning on a specific collection of datasets or even a single task. 
We experimented with fine-tuning on (the training partition of) all the 203 datasets at the same time. 
However, we observe that, if during this fine-tuning we fix the target column to always be the one corresponding to the prediction task to be evaluated, we rapidly encountered overfitting and prediction scores degrade. This might be a similar problem to the scaling law of training data, as also observed by~\citet{tabdpt}. In the present situation, we can avoid overfitting by selecting the target column randomly with the same selection procedure described in Section~\ref{sec:training}. However, as shown in Table~\ref{table:ablation}, the gain is rather limited.

\begin{figure}
    \centering
    \begin{subfigure}[b]{0.475\textwidth}
        \includegraphics[width=\textwidth]{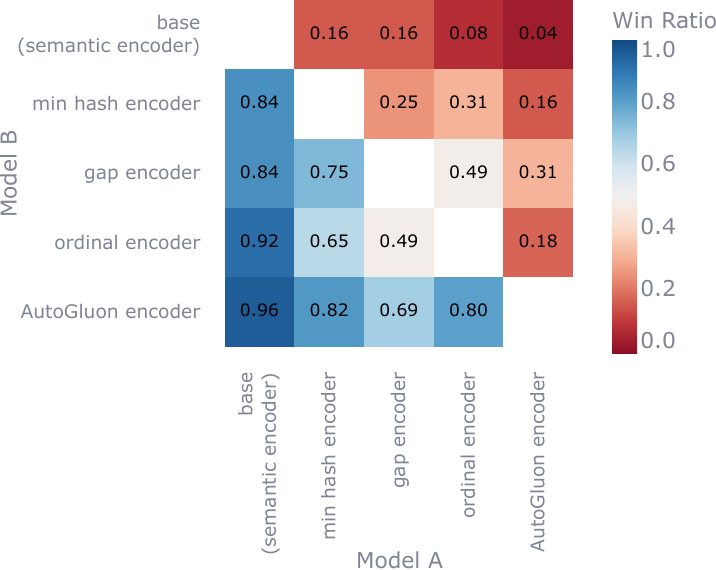}
        \caption{Win ratios with Model A wins over Model B.}
    \end{subfigure}
    \hfill
    \begin{subfigure}[b]{0.475\textwidth}
        \includegraphics[width=\textwidth]{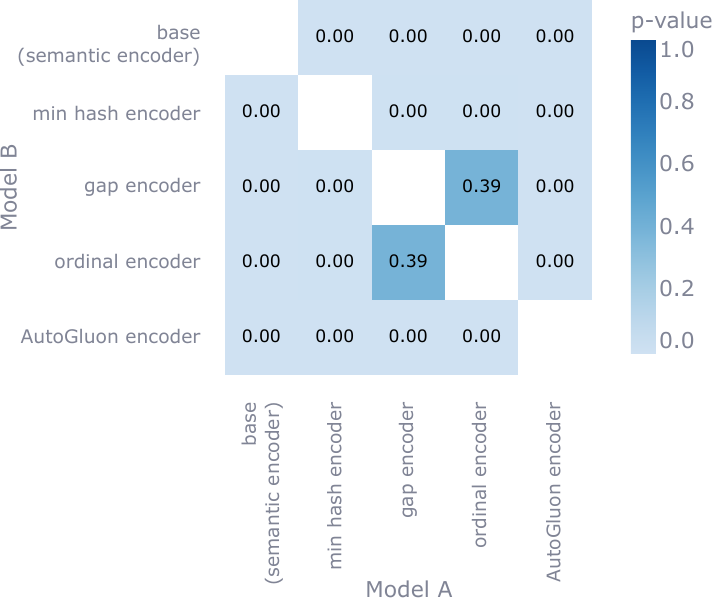}
        \caption{P-values of two-sided Wilcoxon signed-rank test.}
    \end{subfigure}
\caption{Detailed results of the semantic feature ablation, comparing the use of different non-semantic or conventional encoders to the base model using LLM embeddings.}
\label{fig:semantic-ablation-cell-values}
\end{figure}

\begin{figure}
    \centering
    \begin{subfigure}[b]{0.475\textwidth}
        \includegraphics[width=\textwidth]{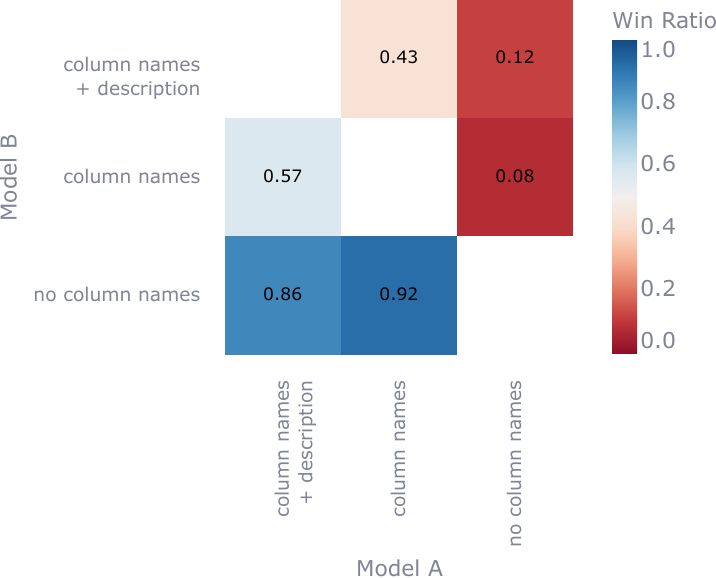}
        \caption{Win ratios with Model A wins over Model B.}
    \end{subfigure}
    \hfill
    \begin{subfigure}[b]{0.475\textwidth}
        \includegraphics[width=\textwidth]{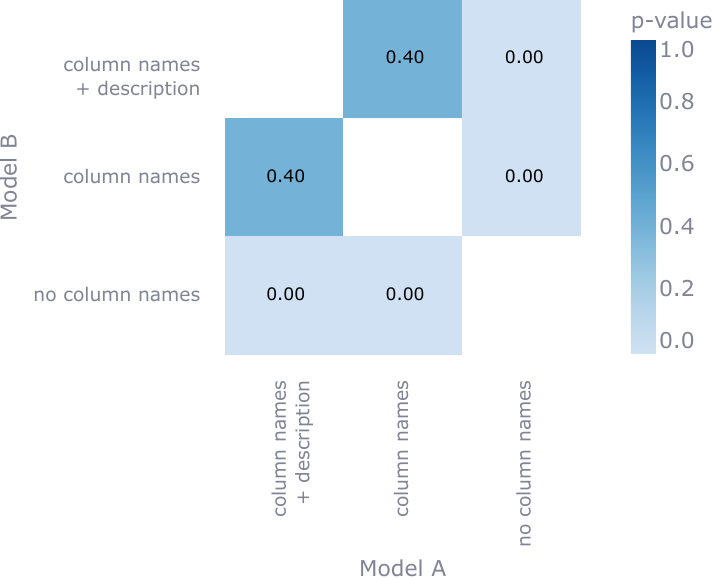}
        \caption{P-values of two-sided Wilcoxon signed-rank test.}
    \end{subfigure}
\caption{Detailed results of the semantic column name ablation, comparing the base model (``column names'') to variations where column names are replaced with non-semantic ones (``no column names'') or enriched with additional column descriptions (``column names'' + description).}
\label{fig:semantic-ablation-column-names}
\end{figure}

\custompar{Semantic embeddings ablation}
In addition to the ablation results previously shown in Section~\ref{sec:ablation}, we depict the win ratios and p-values of a two-sided Wilcoxon signed-rank test for the cell value and column name semantic ablations in the Figure~\ref{fig:semantic-ablation-cell-values} and Figure~\ref{fig:semantic-ablation-column-names}. 

\section {Suplementary Diagrams}

\begin{figure}[H]
    \centering
    \includegraphics[width=1.0\linewidth]{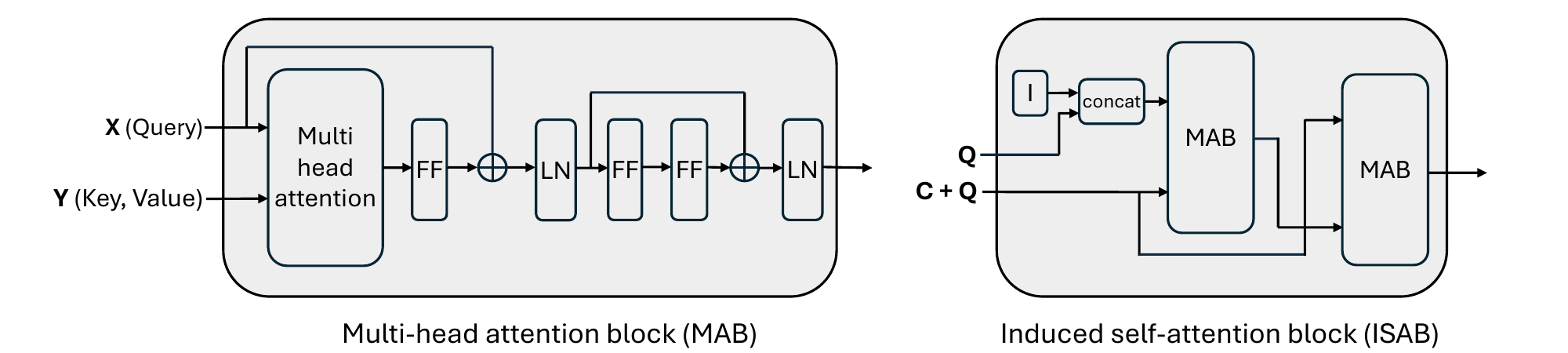}
    \caption{(left) Multi-head attention block (MAB), (right) Induced self-attention block (ISAB) architecture diagram.}
    \label{fig:ISAB_diagram}
\end{figure}

\section{Experimental Setup -- Details}

\subsection{Baselines}\label{app:baselines}

\custompar{Data preprocessor}
Evaluating models across a multitude of datasets can be tricky. Datasets may have inconsistent data type annotations, such as categories represented as strings or categorical data types, covering low- and high-cardinality categorical features, date, time, or datetime instances, free text, boolean values and more. Most models, however, require numerical input to process and handle non-numerical data types or missing values differently. To unify our evaluation, we implemented a configurable default feature encoder built on the \texttt{AutoMLPipelineFeatureGenerator} from AutoGluon~\citep{autogluon} which we found to be very versatile and robust. In particular, the encoder natively handles low- and high-cardinality categorical data, free text (to some extent), as well as datetime encoding. For flexibility and compatibility with a multitude of models, we extended the default implementation to cover the following options that can be adapted to the capabilities of the baseline model at hand:
\begin{itemize}
    \item Whether to convert booleans to string/categorical values.
    \item Whether to \emph{not} encode string/categorical values for models that natively handle them, such as CatBoost.
    \item Whether to scale numerical data via quantile scaling with a normal distribution as target.
    \item Whether to drop constant features.
    \item Whether to impute missing values, extending the standard imputation (using most frequent categories and mean for numerical data) to bools and datetime data types.
\end{itemize}
As default, we choose to convert booleans, encode categoricals via ordinal encoding, scale numerical data, drop constant columns and impute missing values.

Below, we describe for which baselines the default values are changed or when other types of feature encodings are used.

\custompar{TabPFN}
We use the model from the official Python \texttt{tabpfn} package with version 2.1.0 together with the
\texttt{tabpfn-extensions} package version 0.1.0 at commit \texttt{16e0e4f4305a3546eab5be6ebf163ff41bd3843d} of the Git repository\footnote{\url{https://github.com/PriorLabs/tabpfn-extensions.git}} as the PyPi release is not up-to-date.

Naturally, we use \texttt{TabPFNClassifier} and \texttt{TabPFNRegressor} for classification and regression tasks, respectively, using  default parameters for both. In particular, \texttt{TabPFNClassifier} uses an ensemble of 4 and \texttt{TabPFNRegressor} an ensemble of 8 estimators. We combine the classification estimator with the \texttt{ManyClassClassifier} extension with a redundancy factor of 4 to enable classification beyond the native 10-class limit of TabPFN which is required for the evaluation of some of the 203 evaluated datasets.

For datasets larger than the native 10\,k limit of TabPFN, we sample a random 10\,k subset of the training split. This affects 66 out of the 203 evaluated datasets. For datasets with more than the 500 feature limit that TabPFN was trained with, we select a random subsample of 500 features. This affects 12 out of 203 evaluated datasets. While this is not optimal, and post-hoc ensembling as well as a random forest preprocessing is recommended by the authors~\citep{tabpfnv2}, these extensions cannot be combined with the many-class extension required to predict beyond the 10-class limit of the native TabPFN model. Hence, we cannot evaluate TabPFN with the post-hoc ensembling or random forest extension.

As we found the native feature encoder of TabPFN to not work across all evaluated datasets, we use our standard feature encoder (see above), encoding categorical columns, scaling numerical values, dropping constant columns, and imputing missing values. As this procedure should be very similar to the TabPFN-native encoder, we anticipate this deviation to affect the results only insignificantly (if at all).

\custompar{TabICL}
We use the latest model weights \texttt{tabicl-classifier-v1.1-0506.ckpt} from the recent 0.1.2 version of the official \texttt{tabicl} package. This updated variant is an improved checkpoint over the one reported in the original works~\citep{tabicl}. As TabICL solely supports classification tasks, we exclude it from the overall mean rank evaluation.
For encoding, we use our default encoder, but do not scale numericals, do not drop constant values, and do not impute missing ones as it is natively handled by the model.

\custompar{TabDPT}
We use the model from the official GitHub repository\footnote{\url{https://github.com/layer6ai-labs/TabDPT.git}} at the recent 1.1.0 release and \texttt{tabdpt1\_1.pth} model checkpoint.
Naturally, we use \texttt{TabDPTClassifier} and \texttt{TabDPTRegressor} for classification and regression tasks, respectively, using  default parameters for both.
Throughout, we evaluate the model with a (local) context size of 2048 which is the best-performing one in the original works~\citep{tabdpt}. However, evaluation failed for some datasets due to an error in the original code leading to empty predictions for very large datasets in the TabReD benchmark.
Here, we use our default encoder, scaling numericals, dropping constant values, and imputing missing ones.

\custompar{CARTE}
We use the model provided in the official Python \texttt{carte-ai} package with version 0.0.26. 
We use \texttt{CARTEClassifier} and \texttt{CARTERegressor} with default parameters for classification and regression tasks, respectively.
We treat binary classification tasks as 2-class multi-class classification and hence set \texttt{loss=``categorical\_crossentropy''} for the classifacation estimator.
With CARTE, we use our default preprocessor to convert bool values and datetime instances and to impute missing values, but otherwise rely on the \texttt{Table2GraphTransformer} provided in the reference implementation.

\custompar{Pytabkit models} 
We use the \texttt{pytabkit} implementation wrapper for evaluating RealMLP, TabM, XGBoost, LightGBM, and CatBoost. We evaluate all models both in the tuned-defaults (TD) variant proposed by~\citet{realmlp} (except for TabM which only has non-tuned defaults [D]), hyperparameter-optimized (HPO), as well as 5-fold inner cross-validation (CV) ensembled HPO variants (HPO-CV). For all HPO and HPO-CV variants, we use the recently added \texttt{tabarena} search spaces proposed in~\cite{tabarena}.

In particular, for RealMLP (TD), we use \texttt{RealMLP\_TD\_Classifier} and \texttt{RealMLP\_TD\_Regressor} for classification and regression tasks, respectively.
For RealMLP (HPO, HPO-CV), we use \texttt{RealMLP\_HPO\_Classifier} and \texttt{RealMLP\_HPO\_Regressor} for classification and regression tasks, respectively, conducting the default 50 rounds of random search HPO. For the ensembled variant, we use 5-fold inner CV.

For TabM (D), we use \texttt{TabM\_D\_Classifier} and \texttt{TabM\_D\_Regressor} for classification and regression tasks, respectively.
For TabM (HPO, HPO-CV), we use \texttt{TabM\_HPO\_Classifier} and \texttt{TabM\_HPO\_Regressor} for classification and regression tasks, respectively, conducting the default 50 rounds of random search HPO. For the ensembled variant, we use 5-fold inner CV.

For XGBoost (TD), LightGBM (TD), and CatBoost (TD), we use \texttt{XGB\_TD\_Classifier}, \texttt{XGB\_TD\_Regressor}, \texttt{LGBM\_TD\_Classifier}, \texttt{LGBM\_TD\_Regressor}, \texttt{CatBoost\_TD\_Classifier}, and \texttt{CatBoost\_TD\_Regressor} for classification and regression tasks, respectively.
For the HPO-variants, we use the \texttt{HPO\_TPE} versions of the estimators, performing Parzen-tree based HPO with 50 rounds. For the ensembled variant, we use 5-fold inner CV.

CatBoost, LightGBM and XGBoost are evaluated on CPU machines with up to 256\,GB of RAM, whereas RealMLP and TabM are evaluated on H100 GPUs with 96\,GB of VRAM. 

Throughout, we use our default encoder, scaling numericals, dropping constant values, and imputing missing ones. For all models but CatBoost, we perform ordinal encoding of categoricals.

\custompar{Sklearn models}
We use several standard baseline models from \texttt{scikit-learn}~\citep{sklearn}, combining them with the default preprocessor as outlined above. Across all \texttt{scikit-learn} baselines, preprocessing only differs in missing value imputation, depending on the model's capability to handle missing values natively. Throughout, evaluation is performed using \texttt{scikit-learn} v1.5.2.

For the naive predictor, we use the \texttt{DummyClassifier} and \texttt{DummyRegressor} to predict the most frequent, respectively mean value of the train splits as the naive majority baseline.

For the linear predictor, we use the \texttt{LogisticRegression} and \texttt{LinearRegression} for classification and regression tasks, respectively, using default hyperparameters.

For the KNN predictor, we use the \texttt{KNeighborsClassifier} and \texttt{KNeighborsRegressor} for classification and regression tasks, respectively, using default hyperparameters and $k=5$ nearest neighbors.

For the random forest predictor, we use the \texttt{RandomForestClassifier} and \texttt{RandomForestRegressor} for classification and regression tasks, respectively, using default hyperparameters. The model handles missing values natively.

Finally, for the histogram-based gradient boosted tree predictor, we use the \texttt{HistGradientBoostingClassifier} and \texttt{HistGradientBoostingRegressor} for classification and regression tasks, respectively, using default hyperparameters. The model handles missing values natively.

\custompar{Mambular}
We use the official implementation from the \texttt{mambular} PyPi package, version 1.5.1. We use \texttt{MambularClassifier} and \texttt{MambularRegressor} with default parameters. As training ran out of memory for very wide datasets, we randomly sample 500 columns if the dataset exceeds it.

\custompar{AutoGluon}
Throughout, we use AutoGluon v1.2 and its \texttt{TabularPredictor} without custom preprocessing. We use the \texttt{best\_quality} preset and set a per-dataset time limit of 4\,h. Otherwise, parameters are left at their default values. For all datasets, evaluation is executed on a single 16-core machine with 128\,GB of RAM and no GPU.

\subsection{Datasets}\label{app:dataset} 
\begin{figure}
    \centering
    \includegraphics[width=0.9\linewidth]{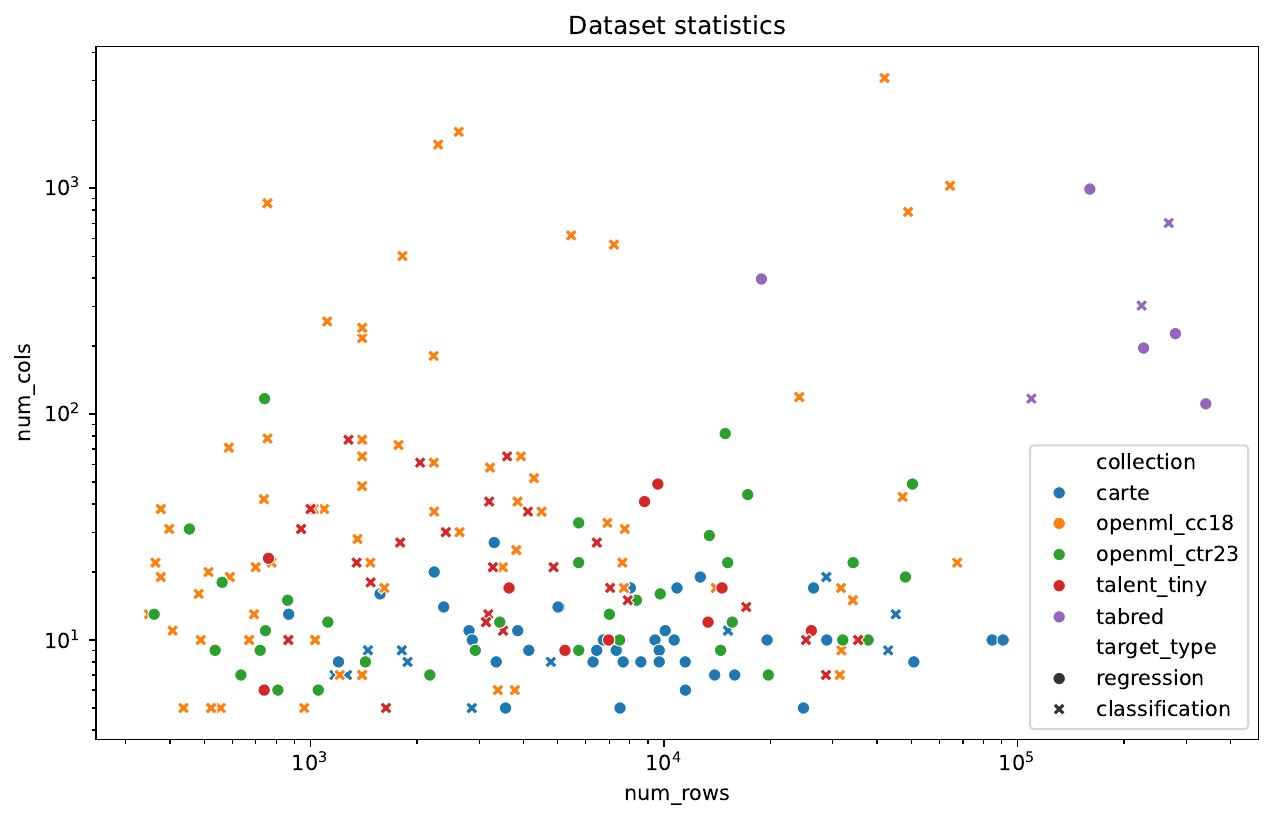}
    \caption{Column and row distribution of the evaluated benchmark datasets.}
    \label{fig:dataset-statistics}
\end{figure}

Full details of all used dataset and benchmarks are provided in Table~\ref{tab:all-dataset-tables}. 
The row and column count statistics are further visualized in Figure~\ref{fig:dataset-statistics}.
Additionally, the per-benchmark statistics regarding the number of tasks as well as the data type proportions are presented in Table~\ref{tab:per-benchmark-data-statistic}. Note that not all types grouped as \texttt{object} are necessarily free text, but that they can also contain categorical or time/date instances that are not correctly cast by pandas by default. However, these are typically picked up correctly by the used data preprocessors.

We extracted all datasets from their original source and performed a custom stratified train-validation-test split with a 70-10-20 ratio. For classification tasks, the target column is used for stratification. For regression tasks, we perform stratification on the binned target column, binning it into 5 quantiles using the \texttt{qcut} method from the \texttt{pandas} library. Otherwise, we do not perform any alterations on the data. Models not using a specific validation procedure are provided with the concatenated train and validation split for training.

\begin{table}
    \centering
    \footnotesize
    \caption{Task and data type statistics of the evaluated benchmarks.}\vspace{2mm}
    \label{tab:per-benchmark-data-statistic}
    \begin{tabular}{lrrrrrrrrr}
    \toprule
     \multirow{2}{*}{Benchmark} & \multicolumn{3}{c}{\textbf{Number of tasks}} & \multicolumn{6}{c}{\textbf{Dtype rates / \%}} \\
    \cmidrule(lr){2-4} \cmidrule(lr){5-10}
     & \multicolumn{1}{c}{total} & \multicolumn{1}{c}{classification} & \multicolumn{1}{c}{regression} & \multicolumn{1}{c}{object} & \multicolumn{1}{c}{category} & \multicolumn{1}{c}{bool} & \multicolumn{1}{c}{float} & \multicolumn{1}{c}{int} & \multicolumn{1}{c}{datetime} \\
    \midrule
    CARTE & 51 & 11 & 40 & 80.74 & 0.00 & 0.00 & 19.04 & 0.22 & 0.00 \\
    OpenML-CC18 & 72 & 72 & 0 & 0.05 & 7.85 & 0.00 & 55.80 & 22.70 & 0.00 \\
    OpenML-CTR23 & 35 & 0 & 35 & 0.95 & 3.91 & 0.00 & 66.19 & 16.91 & 0.00 \\
    TabReD & 8 & 3 & 5 & 0.14 & 0.00 & 0.05 & 89.95 & 9.41 & 0.44 \\
    TALENT-tiny & 37 & 26 & 11 & 12.77 & 0.00 & 0.44 & 63.63 & 23.15 & 0.00 \\
    \bottomrule
    \end{tabular}
\end{table}

\include{dataset_all_tables}

\end{document}

%% file: main_table_small.tex
\begin{tabular}{lrrrrrrrrrrrrrrr}
\toprule
\textbf{Model Name} &
\multicolumn{1}{c}{\textbf{All}} &
\multicolumn{3}{c}{\textbf{CARTE}} &
\multicolumn{2}{c}{\textbf{OML-CC18}} &
\multicolumn{2}{c}{\textbf{OML-CTR23}} &
\multicolumn{3}{c}{\textbf{TabReD}} &
\multicolumn{3}{c}{\textbf{TALENT-Tiny}} \\
\cmidrule(lr){2-2} \cmidrule(lr){3-5} \cmidrule(lr){6-7} \cmidrule(lr){8-9}
\cmidrule(lr){10-12} \cmidrule(lr){13-15}
&
\multicolumn{1}{c}{\textbf{Rank}} &
\multicolumn{1}{c}{\textbf{Rank}} & \multicolumn{1}{c}{\textbf{Acc}} & \multicolumn{1}{c}{\textbf{R$^\textrm{2}$}}   &
\multicolumn{1}{c}{\textbf{Rank}} & \multicolumn{1}{c}{\textbf{Acc}}  &
\multicolumn{1}{c}{\textbf{Rank}} & \multicolumn{1}{c}{\textbf{R$^\textrm{2}$}}  &
\multicolumn{1}{c}{\textbf{Rank}} & \multicolumn{1}{c}{\textbf{Acc}}  & \multicolumn{1}{c}{\textbf{R$^\textrm{2}$}}   &
\multicolumn{1}{c}{\textbf{Rank}} & \multicolumn{1}{c}{\textbf{Acc}}  & \multicolumn{1}{c}{\textbf{R$^\textrm{2}$}}  \\
\midrule
AutoGluon [v1.2] & \NAN & \NAN & 78.7 & 73.8 & \NAN & 88.5 & \NAN & 68.6 & \NAN & 86.0 & 64.6 & \NAN & 87.9 & 73.7\\
\arrayrulecolor{mygrey}\midrule\arrayrulecolor{black}

\textbf{\ours{}} & 2.96 & \bf{1.55} & \bf{76.9} & \bf{72.4} & 3.56 & 86.6 & 3.89 & 73.3 & 4.12 & 85.4 & 63.2 & 2.59 & 87.6 & 76.1\\
LightGBM [HPO, CV] & 2.85 & 3.00 & 73.4 & 67.5 & 2.94 & 87.5 & 2.91 & 68.8 & \bf{1.25} & 86.0 & \bf{64.9} & 2.73 & 86.9 & 74.1\\
CatBoost [HPO, CV] & 2.71 & 3.06 & 75.9 & 68.2 & 2.72 & 87.7 & 2.74 & 68.8 & 2.00 & 86.0 & 64.3 & \bf{2.32} & 87.2 & \bf{76.4}\\
RealMLP [HPO, CV] & \bf{2.50} & 3.20 & 73.4 & 67.5 & \bf{2.35} & \bf{88.3} & \bf{1.89} & 75.0 & 1.88 & 86.0 & 64.6 & 2.54 & \bf{88.3} & \bf{76.4}\\
TabM [HPO, CV] & 2.98 & 3.71 & 73.2 & 66.9 & 2.93 & 87.9 & 2.37 & 74.3 & 3.12 & \bf{86.1} & 64.2 & 2.62 & 87.5 & 75.7\\
XGBoost [HPO, CV] & 3.11 & 3.78 & 73.1 & 66.7 & 2.83 & 87.5 & 3.23 & 70.9 & \bf{1.25} & 86.0 & \bf{64.9} & 3.00 & 86.8 & 73.6\\
TabPFN [v2.1.0] & 3.63 & 5.76 & 72.3 & 65.0 & 2.89 & 87.4 & 3.29 & \bf{75.1} & 4.25 & 85.6 & 63.8 & \bf{2.32} & 87.8 & 74.4\\
TabDPT [v1.1, k=2048] & 4.18 & 5.98 & 72.7 & 65.1 & 3.25 & 87.8 & 3.17 & 73.2 & 6.88 & 83.0 & 60.9 & 3.89 & 86.7 & 74.8\\
HistGradBoost & 5.27 & 6.27 & 72.5 & 64.8 & 4.40 & 86.1 & 7.06 & 65.5 & 2.38 & 85.9 & 63.9 & 4.51 & 86.3 & 67.6\\
Random forest & 6.22 & 8.00 & 71.5 & 63.3 & 5.33 & 85.7 & 6.80 & 67.9 & 6.00 & 85.4 & 60.7 & 5.00 & 85.8 & 70.6\\
Naive & 10.51 & 11.0 & 53.0 & -1.8 & 10.28 & 47.0 & 10.66 & -8.4 & 9.75 & 80.8 & -0.6 & 10.32 & 53.4 & -19.2\\

\arrayrulecolor{mygrey}\midrule\arrayrulecolor{black}
CARTE [v0.0.26] & \NAN & \NAN & 76.1 & 68.5 & \NAN & \NAN & \NAN & \NAN & \NAN & \NAN & \NAN & \NAN & \NAN & \NAN\\
TabICL [v0.1.1]  & \NAN & \NAN & 72.5 & \NAN & \NAN & 88.0 & \NAN & \NAN & \NAN & 85.1 & \NAN & \NAN & 87.4 & \NAN\\
\bottomrule
\end{tabular}

%% file: main_table.tex
\begin{tabular}{lrrrrrrrrrrrrrrr}
\toprule
\textbf{Model Name} &
\multicolumn{1}{c}{\textbf{All}} &
\multicolumn{3}{c}{\textbf{CARTE}} &
\multicolumn{2}{c}{\textbf{OML-CC18}} &
\multicolumn{2}{c}{\textbf{OML-CTR23}} &
\multicolumn{3}{c}{\textbf{TabReD}} &
\multicolumn{3}{c}{\textbf{TALENT-Tiny}} \\
\cmidrule(lr){2-2} \cmidrule(lr){3-5} \cmidrule(lr){6-7} \cmidrule(lr){8-9}
\cmidrule(lr){10-12} \cmidrule(lr){13-15}
&
\multicolumn{1}{c}{\textbf{Rank}} &
\multicolumn{1}{c}{\textbf{Rank}} & \multicolumn{1}{c}{\textbf{Acc}} & \multicolumn{1}{c}{\textbf{R$^\textrm{2}$}}   &
\multicolumn{1}{c}{\textbf{Rank}} & \multicolumn{1}{c}{\textbf{Acc}}  &
\multicolumn{1}{c}{\textbf{Rank}} & \multicolumn{1}{c}{\textbf{R$^\textrm{2}$}}  &
\multicolumn{1}{c}{\textbf{Rank}} & \multicolumn{1}{c}{\textbf{Acc}}  & \multicolumn{1}{c}{\textbf{R$^\textrm{2}$}}   &
\multicolumn{1}{c}{\textbf{Rank}} & \multicolumn{1}{c}{\textbf{Acc}}  & \multicolumn{1}{c}{\textbf{R$^\textrm{2}$}}  \\
\midrule

AutoGluon [v1.2] & \bf{3.34} & \bf{1.92} & \bf{78.7} & \bf{73.8} & 3.42 & \bf{88.5} & 6.17 & 68.6 & 2.25 & 86.0 & 64.6 & \bf{2.70} & 87.9 & 73.7\\
ConTextTab [bagging=8] & 4.70 & 2.18 & 76.9 & 72.4 & 5.65 & 86.6 & 6.74 & 73.3 & 5.38 & 85.4 & 63.2 & 4.24 & 87.6 & 76.1\\
ConTextTab [bagging=1] & 5.42 & 3.22 & 76.4 & 71.5 & 6.51 & 86.3 & 6.91 & 73.1 & 8.50 & 85.3 & 62.6 & 4.27 & 87.6 & 76.0\\
CatBoost [HPO, CV] & 4.10 & 4.82 & 75.9 & 68.2 & 4.25 & 87.7 & 4.29 & 68.8 & 2.88 & 86.0 & 64.3 & 2.89 & 87.2 & \bf{76.4}\\
LightGBM [HPO, CV] & 4.43 & 5.24 & 73.4 & 67.5 & 4.68 & 87.5 & 4.11 & 68.8 & \bf{1.38} & 86.0 & \bf{64.9} & 3.81 & 86.9 & 74.1\\
RealMLP [HPO, CV] & 3.61 & 5.71 & 73.4 & 67.5 & \bf{3.26} & 88.3 & \bf{2.54} & 75.0 & 2.25 & 86.0 & 64.6 & \bf{2.70} & \bf{88.3} & \bf{76.4}\\
TabM [HPO, CV] & 4.17 & 6.08 & 73.2 & 66.9 & 4.35 & 87.9 & 2.63 & 74.3 & 3.62 & \bf{86.1} & 64.2 & 2.76 & 87.5 & 75.7\\
XGBoost [HPO, CV] & 4.66 & 6.12 & 73.1 & 66.7 & 4.03 & 87.5 & 4.60 & 70.9 & \bf{1.38} & 86.0 & \bf{64.9} & 4.62 & 86.8 & 73.6\\
CatBoost [HPO] & 5.13 & 6.16 & 75.4 & 67.1 & 5.04 & 87.2 & 5.31 & 67.4 & 2.25 & 85.9 & 64.3 & 4.35 & 86.5 & 73.3\\
LightGBM [HPO] & 6.26 & 7.63 & 72.9 & 66.3 & 6.29 & 87.2 & 6.09 & 67.8 & 2.25 & 85.9 & 64.3 & 5.35 & 86.5 & 73.2\\
XGBoost [HPO] & 6.61 & 8.41 & 72.7 & 65.8 & 7.00 & 86.7 & 5.66 & 69.2 & \bf{1.38} & 85.8 & 64.6 & 5.41 & 86.4 & 74.0\\
CatBoost [TD] & 6.83 & 8.61 & 75.4 & 65.8 & 6.04 & 86.9 & 8.57 & 70.7 & 3.00 & 85.9 & 63.6 & 5.08 & 86.2 & 74.2\\
TabM [HPO] & 5.42 & 9.29 & 72.6 & 65.6 & 4.54 & 87.6 & 4.46 & 73.3 & 3.38 & \bf{86.1} & 64.6 & 3.16 & 87.3 & 75.5\\
TabDPT [v1.1, k=2048] & 6.83 & 9.84 & 72.7 & 65.1 & 5.01 & 87.8 & 5.69 & 73.2 & 15.12 & 83.0 & 60.9 & 5.51 & 86.7 & 74.8\\
TabPFN [v2.1.0] & 5.81 & 10.08 & 72.3 & 65.0 & 4.85 & 87.4 & 4.74 & \bf{75.1} & 5.25 & 85.6 & 63.8 & 2.92 & 87.8 & 74.4\\
HistGradientBoosting & 8.92 & 10.53 & 72.5 & 64.8 & 8.01 & 86.1 & 11.37 & 65.5 & 3.25 & 85.9 & 63.9 & 7.38 & 86.3 & 67.6\\
LightGBM [TD] & 7.80 & 10.75 & 72.7 & 64.9 & 6.49 & 86.4 & 9.23 & 67.5 & 3.25 & 85.9 & 63.5 & 5.95 & 86.0 & 72.9\\
XGBoost [TD] & 8.03 & 12.18 & 72.5 & 64.2 & 6.03 & 86.7 & 9.34 & 68.7 & 3.75 & 85.6 & 62.6 & 5.89 & 86.2 & 72.7\\
RealMLP [HPO] & 7.10 & 12.49 & 71.6 & 64.6 & 4.89 & 87.5 & 6.14 & 70.6 & 3.50 & 85.9 & 63.7 & 5.68 & 86.5 & 76.3\\
Random forest & 11.01 & 14.08 & 71.5 & 63.3 & 9.60 & 85.7 & 11.54 & 67.9 & 10.12 & 85.4 & 60.7 & 9.22 & 85.8 & 70.6\\
TabM & 9.42 & 16.96 & 70.1 & 60.5 & 7.14 & 86.7 & 6.60 & 69.7 & 10.25 & 85.3 & 62.2 & 5.97 & 86.1 & 74.3\\
RealMLP [TD] & 8.86 & 17.96 & 69.4 & 60.1 & 5.39 & 87.1 & 7.06 & 71.0 & 3.00 & 85.9 & 64.0 & 6.03 & 86.2 & 74.7\\
KNN [k=5] & 20.04 & 22.9 & 65.5 & 34.3 & 17.94 & 81.7 & 20.46 & 55.1 & 21.62 & 78.7 & -15.0 & 19.46 & 80.3 & 60.0\\
Linear & 19.52 & 23.25 & 62.7 & 22.8 & 16.39 & 80.9 & 21.63 & 46.2 & 20.12 & 80.8 & 21.1 & 18.35 & 80.5 & 41.3\\
Naive & 23.33 & 24.82 & 53.0 & -1.8 & 22.43 & 47.0 & 24.11 & -8.4 & 20.75 & 80.8 & -0.6 & 22.84 & 53.4 & -19.2\\

\arrayrulecolor{mygrey}\midrule\arrayrulecolor{black}
CARTE [v0.0.26]  & \NAN & \NAN & 76.1 & 68.5 & \NAN & \NAN & \NAN & \NAN & \NAN & \NAN & \NAN & \NAN & 84.4 & 71.1\\
TabICL  [v0.1.1] & \NAN & \NAN & 72.5 & \NAN & \NAN & 88.0 & \NAN & \NAN & \NAN & 85.1 & \NAN & \NAN & 87.4 & \NAN\\
Mambular [v1.5.1] & \NAN & \NAN & 70.1 & 52.8 & \NAN & \NAN & \NAN & \NAN & \NAN & \NAN & \NAN & \NAN & 83.6 & 53.9\\

\bottomrule
\end{tabular}

%% file: dataset_all_tables.tex
\begin{longtable}{llrrrr}
\caption{Details of all used benchmark datasets.}\label{tab:all-dataset-tables} \\
\toprule
Benchmark & Table & Num. rows & Num. cols & Target type \\
\midrule
\endfirsthead \\
\toprule
Collection & Table & Num. rows & Num. cols & Target type \\
\midrule
\endhead
\midrule
\multicolumn{5}{r}{Continued on next page} \\
\midrule
\endfoot
\bottomrule
\endlastfoot
CARTE & anime\_planet & 11512 & 11 & regression \\
CARTE & babies\_r\_us & 4068 & 5 & regression \\
CARTE & beer\_ratings & 2557 & 20 & regression \\
CARTE & bikedekho & 3828 & 8 & regression \\
CARTE & bikewale & 7193 & 8 & regression \\
CARTE & buy\_buy\_baby & 8574 & 5 & regression \\
CARTE & cardekho & 30250 & 17 & regression \\
CARTE & chocolate\_bar\_ratings & 2070 & 9 & classification \\
CARTE & clear\_corpus & 3779 & 27 & regression \\
CARTE & coffee\_ratings & 1661 & 9 & classification \\
CARTE & company\_employees & 8752 & 8 & regression \\
CARTE & employee\_remuneration & 28316 & 5 & regression \\
CARTE & employee\_salaries & 7368 & 9 & regression \\
CARTE & fifa22\_players & 14468 & 19 & regression \\
CARTE & filmtv\_movies & 32964 & 10 & regression \\
CARTE & journal\_jcr & 7692 & 10 & regression \\
CARTE & journal\_sjr & 22344 & 10 & regression \\
CARTE & jp\_anime & 12428 & 17 & regression \\
CARTE & k\_drama & 991 & 13 & regression \\
CARTE & michelin & 5465 & 8 & classification \\
CARTE & mlds\_salaries & 8364 & 9 & regression \\
CARTE & movies & 5779 & 14 & regression \\
CARTE & museums & 9173 & 17 & regression \\
CARTE & mydramalist & 2720 & 14 & regression \\
CARTE & nba\_draft & 1335 & 7 & classification \\
CARTE & prescription\_drugs & 1371 & 8 & regression \\
CARTE & ramen\_ratings & 3267 & 5 & classification \\
CARTE & roger\_ebert & 2149 & 8 & classification \\
CARTE & rotten\_tomatoes & 5726 & 14 & regression \\
CARTE & spotify & 32879 & 19 & classification \\
CARTE & us\_accidents\_counts & 18098 & 7 & regression \\
CARTE & us\_accidents\_severity & 17324 & 11 & classification \\
CARTE & us\_presidential & 15885 & 7 & regression \\
CARTE & used\_cars\_24 & 4734 & 9 & regression \\
CARTE & used\_cars\_benz\_italy & 13112 & 8 & regression \\
CARTE & used\_cars\_dot\_com & 3207 & 11 & regression \\
CARTE & used\_cars\_pakistan & 58124 & 8 & regression \\
CARTE & used\_cars\_saudi\_arabia & 4405 & 11 & regression \\
CARTE & videogame\_sales & 13128 & 6 & regression \\
CARTE & whisky & 1449 & 7 & classification \\
CARTE & wikiliq\_beer & 10768 & 10 & regression \\
CARTE & wikiliq\_spirit & 9820 & 8 & regression \\
CARTE & wina\_pl & 1797 & 16 & regression \\
CARTE & wine\_dot\_com\_prices & 12203 & 10 & regression \\
CARTE & wine\_dot\_com\_ratings & 3276 & 10 & regression \\
CARTE & wine\_enthusiasts\_prices & 96780 & 10 & regression \\
CARTE & wine\_enthusiasts\_ratings & 103976 & 10 & regression \\
CARTE & wine\_vivino\_price & 11067 & 8 & regression \\
CARTE & wine\_vivino\_rating & 11067 & 9 & regression \\
CARTE & yelp & 51692 & 13 & classification \\
CARTE & zomato & 49152 & 9 & classification \\
OpenML-CC18 & adult & 39073 & 15 & classification \\
OpenML-CC18 & analcatdata\_authorship & 672 & 71 & classification \\
OpenML-CC18 & analcatdata\_dmft & 637 & 5 & classification \\
OpenML-CC18 & balance\_scale & 500 & 5 & classification \\
OpenML-CC18 & bank\_marketing & 36168 & 17 & classification \\
OpenML-CC18 & banknote\_authentication & 1097 & 5 & classification \\
OpenML-CC18 & bioresponse & 3000 & 1777 & classification \\
OpenML-CC18 & blood\_transfusion\_service\_center & 598 & 5 & classification \\
OpenML-CC18 & breast\_w & 559 & 10 & classification \\
OpenML-CC18 & car & 1382 & 7 & classification \\
OpenML-CC18 & churn & 4000 & 21 & classification \\
OpenML-CC18 & cifar\_10 & 48000 & 3073 & classification \\
OpenML-CC18 & climate\_model\_simulation\_crashes & 432 & 19 & classification \\
OpenML-CC18 & cmc & 1178 & 10 & classification \\
OpenML-CC18 & cnae\_9 & 864 & 857 & classification \\
OpenML-CC18 & connect\_4 & 54045 & 43 & classification \\
OpenML-CC18 & credit\_approval & 552 & 16 & classification \\
OpenML-CC18 & credit\_g & 800 & 21 & classification \\
OpenML-CC18 & cylinder\_bands & 432 & 38 & classification \\
OpenML-CC18 & devnagari\_script & 73600 & 1025 & classification \\
OpenML-CC18 & diabetes & 614 & 9 & classification \\
OpenML-CC18 & dna & 2548 & 181 & classification \\
OpenML-CC18 & dresses\_sales & 400 & 13 & classification \\
OpenML-CC18 & electricity & 36249 & 9 & classification \\
OpenML-CC18 & eucalyptus & 588 & 20 & classification \\
OpenML-CC18 & fashion\_mnist & 56000 & 785 & classification \\
OpenML-CC18 & first\_order\_theorem\_proving & 4894 & 52 & classification \\
OpenML-CC18 & gesturephasesegmentationprocessed & 7898 & 33 & classification \\
OpenML-CC18 & har & 8239 & 562 & classification \\
OpenML-CC18 & ilpd & 466 & 11 & classification \\
OpenML-CC18 & internet\_advertisements & 2623 & 1559 & classification \\
OpenML-CC18 & isolet & 6237 & 618 & classification \\
OpenML-CC18 & jm1 & 8708 & 22 & classification \\
OpenML-CC18 & jungle\_chess\_2pcs\_raw\_endgame\_complete & 35855 & 7 & classification \\
OpenML-CC18 & kc1 & 1687 & 22 & classification \\
OpenML-CC18 & kc2 & 417 & 22 & classification \\
OpenML-CC18 & kr\_vs\_kp & 2556 & 37 & classification \\
OpenML-CC18 & letter & 16000 & 17 & classification \\
OpenML-CC18 & madelon & 2080 & 501 & classification \\
OpenML-CC18 & mfeat\_factors & 1600 & 217 & classification \\
OpenML-CC18 & mfeat\_fourier & 1600 & 77 & classification \\
OpenML-CC18 & mfeat\_karhunen & 1600 & 65 & classification \\
OpenML-CC18 & mfeat\_morphological & 1600 & 7 & classification \\
OpenML-CC18 & mfeat\_pixel & 1600 & 241 & classification \\
OpenML-CC18 & mfeat\_zernike & 1600 & 48 & classification \\
OpenML-CC18 & miceprotein & 864 & 78 & classification \\
OpenML-CC18 & mnist\_784 & 56000 & 785 & classification \\
OpenML-CC18 & nomao & 27572 & 119 & classification \\
OpenML-CC18 & numerai28\_6 & 77056 & 22 & classification \\
OpenML-CC18 & optdigits & 4496 & 65 & classification \\
OpenML-CC18 & ozone\_level\_8hr & 2027 & 73 & classification \\
OpenML-CC18 & pc1 & 887 & 22 & classification \\
OpenML-CC18 & pc3 & 1250 & 38 & classification \\
OpenML-CC18 & pc4 & 1166 & 38 & classification \\
OpenML-CC18 & pendigits & 8793 & 17 & classification \\
OpenML-CC18 & phishingwebsites & 8844 & 31 & classification \\
OpenML-CC18 & phoneme & 4323 & 6 & classification \\
OpenML-CC18 & qsar\_biodeg & 844 & 42 & classification \\
OpenML-CC18 & satimage & 5144 & 37 & classification \\
OpenML-CC18 & segment & 1848 & 17 & classification \\
OpenML-CC18 & semeion & 1274 & 257 & classification \\
OpenML-CC18 & sick & 3017 & 30 & classification \\
OpenML-CC18 & spambase & 3680 & 58 & classification \\
OpenML-CC18 & splice & 2552 & 61 & classification \\
OpenML-CC18 & steel\_plates\_fault & 1552 & 28 & classification \\
OpenML-CC18 & texture & 4400 & 41 & classification \\
OpenML-CC18 & tic\_tac\_toe & 766 & 10 & classification \\
OpenML-CC18 & vehicle & 676 & 19 & classification \\
OpenML-CC18 & vowel & 792 & 13 & classification \\
OpenML-CC18 & wall\_robot\_navigation & 4364 & 25 & classification \\
OpenML-CC18 & wdbc & 455 & 31 & classification \\
OpenML-CC18 & wilt & 3871 & 6 & classification \\
OpenML-CTR23 & abalone & 3341 & 9 & regression \\
OpenML-CTR23 & airfoil\_self\_noise & 1202 & 6 & regression \\
OpenML-CTR23 & auction\_verification & 1634 & 8 & regression \\
OpenML-CTR23 & brazilian\_houses & 8553 & 10 & regression \\
OpenML-CTR23 & california\_housing & 16512 & 9 & regression \\
OpenML-CTR23 & cars & 643 & 18 & regression \\
OpenML-CTR23 & concrete\_compressive\_strength & 824 & 9 & regression \\
OpenML-CTR23 & cps88wages & 22524 & 7 & regression \\
OpenML-CTR23 & cpu\_activity & 6553 & 22 & regression \\
OpenML-CTR23 & diamonds & 43152 & 10 & regression \\
OpenML-CTR23 & energy\_efficiency & 614 & 9 & regression \\
OpenML-CTR23 & fifa & 15342 & 29 & regression \\
OpenML-CTR23 & forest\_fires & 413 & 13 & regression \\
OpenML-CTR23 & fps\_benchmark & 19699 & 44 & regression \\
OpenML-CTR23 & geographical\_origin\_of\_music & 847 & 117 & regression \\
OpenML-CTR23 & grid\_stability & 8000 & 13 & regression \\
OpenML-CTR23 & health\_insurance & 17817 & 12 & regression \\
OpenML-CTR23 & kin8nm & 6553 & 9 & regression \\
OpenML-CTR23 & kings\_county & 17290 & 22 & regression \\
OpenML-CTR23 & miami\_housing & 11145 & 16 & regression \\
OpenML-CTR23 & moneyball & 985 & 15 & regression \\
OpenML-CTR23 & naval\_propulsion\_plant & 9547 & 15 & regression \\
OpenML-CTR23 & physiochemical\_protein & 36584 & 10 & regression \\
OpenML-CTR23 & pumadyn32nh & 6553 & 33 & regression \\
OpenML-CTR23 & qsar\_fish\_toxicity & 726 & 7 & regression \\
OpenML-CTR23 & red\_wine & 1279 & 12 & regression \\
OpenML-CTR23 & sarcos & 39146 & 22 & regression \\
OpenML-CTR23 & socmob & 924 & 6 & regression \\
OpenML-CTR23 & solar\_flare & 852 & 11 & regression \\
OpenML-CTR23 & space\_ga & 2485 & 7 & regression \\
OpenML-CTR23 & student\_performance\_por & 519 & 31 & regression \\
OpenML-CTR23 & superconductivity & 17010 & 82 & regression \\
OpenML-CTR23 & video\_transcoding & 55027 & 19 & regression \\
OpenML-CTR23 & wave\_energy & 57600 & 49 & regression \\
OpenML-CTR23 & white\_wine & 3918 & 12 & regression \\
TabReD & acquire\_valued\_shoppers\_challenge & 133602 & 117 & classification \\
TabReD & cooking\_time & 278338 & 196 & regression \\
TabReD & delivery\_eta & 313589 & 227 & regression \\
TabReD & home\_credit\_credit\_risk\_model\_stability & 325663 & 701 & classification \\
TabReD & homesite\_quote\_conversion & 244458 & 302 & classification \\
TabReD & maps\_routing & 219994 & 991 & regression \\
TabReD & sberbank\_russian\_housing\_market & 23674 & 396 & regression \\
TabReD & TabReD\_weather & 382955 & 111 & regression \\
TALENT-tiny & ailerons & 11000 & 41 & regression \\
TALENT-tiny & breast\_w & 31492 & 10 & classification \\
TALENT-tiny & cmc & 44236 & 10 & classification \\
TALENT-tiny & dis & 3017 & 30 & classification \\
TALENT-tiny & eye\_movements\_bin & 6086 & 21 & classification \\
TALENT-tiny & fried & 32614 & 11 & regression \\
TALENT-tiny & house\_16h\_reg & 18227 & 17 & regression \\
TALENT-tiny & ibm\_employee\_performance & 1176 & 31 & classification \\
TALENT-tiny & jungle\_chess\_2pcs\_raw\_endgame\_complete & 35855 & 7 & classification \\
TALENT-tiny & kaggle\_bike\_sharing\_demand\_challange & 8708 & 10 & regression \\
TALENT-tiny & kc1 & 1687 & 22 & classification \\
TALENT-tiny & kin8nm & 6553 & 9 & regression \\
TALENT-tiny & law\_school\_admission\_bianry & 16640 & 12 & regression \\
TALENT-tiny & mfeat\_fourier & 1600 & 77 & classification \\
TALENT-tiny & mv & 32614 & 11 & regression \\
TALENT-tiny & okcupid\_stem & 21341 & 14 & classification \\
TALENT-tiny & online\_shoppers & 9864 & 15 & classification \\
TALENT-tiny & optdigits & 4496 & 65 & classification \\
TALENT-tiny & page\_blocks & 4378 & 11 & classification \\
TALENT-tiny & pc3 & 1250 & 38 & classification \\
TALENT-tiny & pendigits & 8793 & 17 & classification \\
TALENT-tiny & pol & 8065 & 27 & classification \\
TALENT-tiny & pol\_reg & 12000 & 49 & regression \\
TALENT-tiny & rl & 3976 & 13 & classification \\
TALENT-tiny & satimage & 5144 & 37 & classification \\
TALENT-tiny & segment & 1848 & 18 & classification \\
TALENT-tiny & socmob & 924 & 6 & regression \\
TALENT-tiny & splice & 2552 & 61 & classification \\
TALENT-tiny & sylvine & 4099 & 21 & classification \\
TALENT-tiny & thyroid\_dis & 2240 & 27 & classification \\
TALENT-tiny & tic\_tac\_toe & 31492 & 10 & classification \\
TALENT-tiny & vulnonevul & 4553 & 17 & regression \\
TALENT-tiny & waterstress & 950 & 23 & regression \\
TALENT-tiny & waveform\_5000 & 4000 & 41 & classification \\
TALENT-tiny & website\_phishing & 1082 & 10 & classification \\
TALENT-tiny & wine & 2043 & 5 & classification \\
TALENT-tiny & wine\_quality\_white & 3918 & 12 & classification \\
\end{longtable}